\pdfoutput=1

\documentclass[11pt]{article}

\usepackage{ACL}

\usepackage{times}
\usepackage{latexsym}

\usepackage[T1]{fontenc}

\usepackage[utf8]{inputenc}

\usepackage{microtype}

\usepackage{inconsolata}

\usepackage{inconsolata}
\usepackage[most]{tcolorbox}
\usepackage{cleveref}
\usepackage{mathrsfs}
\usepackage{algorithm}
\usepackage{algpseudocode}
\usepackage{caption}
\usepackage{array}
\usepackage{booktabs}
\usepackage{tabularx}
\newcolumntype{C}[1]{>{\centering\arraybackslash}p{#1}}
\usepackage{multirow}
\usepackage{enumitem}
\usepackage{todonotes}
\usepackage{tikz, xcolor}
\usepackage{amsmath}
\usepackage[table]{xcolor}
\definecolor{lightgreen}{RGB}{240, 255, 240} 

\title{Diverse, not Short: A Length-Controlled 
Data Selection Strategy
for Improving Response Diversity of Language Models}

\author{
  Vijeta Deshpande\textsuperscript{1} \quad
  Debasmita Ghose\textsuperscript{2} \quad
  John D. Patterson\textsuperscript{3} \\
  \textbf{Roger Beaty\textsuperscript{3}} \quad
  \textbf{Anna Rumshisky\textsuperscript{1,4}} \\
  \textsuperscript{1}University of Massachusetts Lowell \quad
  \textsuperscript{2}Yale University \\
  \textsuperscript{3}Pennsylvania State University \quad
  \textsuperscript{4}Amazon AGI \\
  \texttt{vijeta\_deshpande@student.uml.edu}
}

\begin{document}
\maketitle
\begin{abstract}

Diverse language model responses are crucial for creative generation, open-ended tasks, and self-improvement training. We show that common diversity metrics, and even reward models used for preference optimization, systematically bias models toward shorter outputs, limiting expressiveness. To address this, we introduce Diverse, not Short (Diverse-NS), a length-controlled 
data selection strategy
that improves response diversity while maintaining length parity. By generating and filtering preference data that balances diversity, quality, and length, Diverse-NS enables effective training using only 3,000 preference pairs. Applied to LLaMA-3.1-8B and the Olmo-2 family, Diverse-NS substantially enhances lexical and semantic diversity. We show consistent improvement in diversity with minor reduction or gains in response quality on four creative generation tasks: Divergent Associations, Persona Generation, Alternate Uses, and Creative Writing. Surprisingly, experiments with the Olmo-2 model family (7B, and 13B) show that smaller models like Olmo-2-7B can serve as effective “diversity teachers” for larger models. By explicitly addressing length bias, our method efficiently pushes models toward more diverse and expressive outputs\footnotemark. 
\end{abstract}
\footnotetext{
Code: \url{https://github.com/text-machine-lab/diverse-not-short},
Dataset: \url{https://huggingface.co/datasets/text-machine-lab/diverse-not-short}
}

\section{Introduction}



Alignment has played a key role in making large language models (LLMs) broadly useful, controllable, and safe for real-world applications \citep{schulman2017proximal, bai2022training, dai2023safe, ouyang2022training, longpre2023flan}. As a form of post-training, it typically involves a combination of instruction tuning \citep{longpre2023flan, peng2023instruction, ouyang2022training} and preference optimization \citep{schulman2017proximal, ouyang2022training, rafailov2023direct}, enabling models to follow human instructions and generate responses that are helpful, harmless, and honest \citep{bai2022training, dai2023safe}. However, alignment comes at a cost: several studies have found that alignment 
can significantly reduce the diversity of model outputs
\citep{kirk2023understanding, doshi2024generative, padmakumar2023does, anderson2024homogenization, shaib2024detection, west2025base}.

This decrease in diversity has important consequences. When humans collaborate with aligned models, the content they produce tends to be less original and less varied \citep{doshi2024generative, padmakumar2023does}. At scale, this reduction in diversity can hinder creative ideation and increase output homogeneity \citep{anderson2024homogenization, xu2024echoes}. Beyond creativity, reduced diversity of generated text has a direct impact on the continued 
improvement of LLMs,
with only limited benefits of reduced diversity \citep{deshpande2023honey, muckatira2024emergent}. 
Recent studies have shown that repeatedly training models on their own aligned outputs can lead to a consistent decline in diversity, eventually resulting in model collapse \citep{shumailov2023curse, guo2023curious, seddik2024bad}. 

Despite these challenges, alignment remains essential. The question, then, is not whether to align, but how to preserve or recover the output diversity of aligned models.
In this work, we ask: \textit{Can we increase the response diversity of aligned models while retaining the 
the response quality?}
%

Prior work has explored a range of strategies to improve output diversity of aligned language models, including methods based on prompting, sampling, and targeted training procedures  \citep{lu2024benchmarking, zhang2020trading, tian2023macgyver, li2024entropic, li2025preserving, lanchantin2025diverse, chung2025modifying, qin2025dive}. Sampling techniques such as temperature, top-$p$, and top-$k$ have been shown to increase diversity, though often at the cost of reduced quality \citep{zhang2020trading}. Sequential prompting strategies are also helpful in increasing response diversity \citep{lu2024benchmarking, tian2023macgyver}. However, the computational cost scales rapidly with 
more discussion turns due to increasing context length.
Training approaches have introduced explicit diversity objectives \citep{li2025preserving, chung2025modifying, cideron2024diversity} and entropy regularization \citep{li2024entropic} to encourage more varied outputs. Self-learning methods, where the model generates its own training data, have also been used to promote diversity \citep{tian2024toward, lanchantin2025diverse, qin2025dive}. 

%
However, one critical confound, text length, has received little scrutiny in recent work. Widely used diversity metrics are length-sensitive and consistently assign higher scores to shorter passages \citep{covington2010cutting, mccarthy2010mtld, shaib2024standardizing}. While this bias is less problematic in structured generation tasks, optimizing these metrics can reduce expressiveness in open-ended writing, which thrives on depth and nuance, thereby undermining the very creativity they are meant to cultivate. But even though optimizing length-sensitive metrics can clearly backfire, the role of length in both measuring and improving diversity has been largely overlooked. Our work aims to close this gap.

To address this overlooked confounding factor, we propose \textit{Diverse, not Short} (Diverse-NS), a length-controlled 
data selection strategy
that counteracts the hidden brevity bias in standard diversity metrics and improves diversity in both structured and free-form generation.
The framework first uses sequential prompting to elicit more diverse responses, followed by preference pair curation that improve both diversity and quality while maintaining comparable response lengths (within $\pm 5$ words). Using these preference pairs, we apply Direct Preference Optimization (DPO) \citep{rafailov2023direct} to improve the response diversity of the base model.
Our key contributions are:
\begin{enumerate}[itemsep=0pt, topsep=0pt, partopsep=0pt, leftmargin=*]
\item \textbf{Diverse-NS:} A length-controlled 
data selection strategy
that significantly improves the response diversity of Llama-3.1-8B and Olmo-2-7B using only $3$k preference pairs.

\item \textbf{Diverse-NS-Lite:} A computationally efficient variant that achieves comparable performance to Diverse-NS while significantly reducing the data filtering cost.

\item \textbf{Small-to-large transfer:} We highlight the potential of smaller models to serve as effective “diversity teachers” for larger variants, enabling low-cost diversity alignment.

\item \textbf{Length-controlled diversity evaluation:} We introduce \textit{Diversity Decile}, a diversity metric that adjusts for text length.

\item \textbf{Dataset:} We release a high-quality dataset of $6$k preference pairs generated from Llama-3.1-8B and Olmo-2-7B to support future research on length-aware diversity alignment.
\end{enumerate}

\section{Related Work}

\paragraph{Increasing Diversity without Training.} \citet{zhang2020trading, chung2023increasing}, shows that common sampling methods such as temperature, top-p, top-k, are comparable in terms of increasing the diversity but, increasing diversity often comes at the price of reduced quality. For curating a generic large-scale dataset, prompting methods can boost topical, stylistic, and formatting diversity \citep{li2023textbooks, chen2024diversity, cosmopedia2024, ge2024scaling}. Conversely, for more task-specific datasets, sequential prompting can elicit diverse responses \cite{lu2024benchmarking, tian2023macgyver, qin2025dive}.

\paragraph{Increasing Diversity with Training.} Augmenting method-specific objective functions with elements that directly maximize diversity has been successful in increasing response diversity \cite{li2024entropic, chung2025modifying, li2015diversity, li2025preserving}. The other approach gaining more attention in recent studies is to adopt a three-step procedure: generate diverse data, filter data for improving quality, and fine-tune LLM on the filtered data \citep{lanchantin2025diverse, chung2025modifying, qin2025dive}. This approach has been successful in task-specific alignment, but more generic self-training has still seen limited success \citep{li2023textbooks, cosmopedia2024, shumailov2023curse, guo2023curious, herel2024collapse, seddik2024bad}. Our work is closest to the task-specific alignment studies in the self-learning framework 
\cite{lanchantin2025diverse, qin2025dive}.

\paragraph{Diversity Evaluation.} Evaluation of diversity is challenging for a few reasons: length bias \cite{mccarthy2010mtld, covington2010cutting, mass1972zusammenhang, johnson2023divergent, deshpande2025penalty}, relative difficulty in achieving substantial agreement between humans \citep{chakrabarty2023creativity, chakrabarty2024can, gomez2023confederacy}, and inconsistent human preferences \cite{evans2016learning}. Despite the challenges, many studies have highlighted the compromised diversity of synthetic or human-LLM collaborative text \cite{shaib2024detection, shaib2024standardizing, salkar2022self, padmakumar2023does, guo2023curious, kirk2023understanding, doshi2024generative, anderson2024homogenization}. So, we present 
\emph{Diverse-NS}, to increase the response diversity and propose a metric, \emph{Diversity Decile}, to measure diversity in a length-controlled way. 

\section{Preliminaries}

Self-learning, also known as self-training, is a semi-supervised approach involving three main steps: data generation (pseudo-labeling), data filtering, and model learning \citep{lee2013pseudo, amini2025self}. In our setup, data generation involves sampling text from a language model in response to story-writing prompts. This is followed by filtering, where we construct high-quality preference pairs—two continuations for the same prompt, with one preferred over the other. We refer to the preferred continuation as the ``chosen'' 
and the other as the ``rejected''.
Using this preference dataset, we apply Direct Preference Optimization (DPO) \citep{rafailov2023direct} to train the model to favor the chosen responses.

\section{Data}
\label{section:data}

We describe data generation and filtering pipeline designed to elicit diverse model responses for downstream preference tuning. The pipeline first generates candidate stories using a sequential prompting strategy, then filters the pool of generated responses to form preference pairs suitable for Direct Preference Optimization (DPO) training \citep{rafailov2023direct}. The preference pairs are formed to maximize the diversity and quality gain while maintaining the same length for "chosen" and "rejected" samples. 

\subsection{Data Generation}
\label{sec:data_gem}

\paragraph{Task Setup.}
We focus on a creative writing task to build the dataset for preference learning. The goal is to generate short stories (five sentences) that must include three words specified in the prompt. This task has been extensively validated in studies of human creativity \citep{prabhakaran_thin_2014}. To create a diverse set of prompts, we first curated a list of 300 unique words, $W_u$\footnote{A manually curated list of 20 words was extended using GPT-4o and Claude-3.7.}. 
For generating short stories from LMs, we create prompts by randomly sampling three-word sets from $W_u$.

\paragraph{Sequential Prompting.} 
\label{sec:sequential_prompting}
Given the task setup, we create $1$k story writing prompts, with $1$k unique three-word sets. The exact prompt is provided in \Cref{sec:app:prompts_data_gen}.
We initially sampled $10$k stories (10 per prompt) using a temperature of $1.0$ from each of the following LMs: Llama-8B and Olmo-7B \citep{grattafiori2024llama, olmo20242}. Within the sampled stories, we extracted the repeating Part-Of-Speech (POS) bigrams and found that the start of the story is highly likely to have repetitions across different prompts (refer to \Cref{tab:app:repeating_patterns}). To overcome these repetitions, we performed a second inference call to re-draft the story with additional constraints, an approach similar to \textit{Denial Prompting} presented by \citet{lu2024benchmarking} (refer to \Cref{sec:app:prompts_data_gen} exact prompt). In our case, unlike \citet{lu2024benchmarking}, the constraints we use are specifically targeted to elicit a more diverse response from the model while maintaining the same (or comparable) length. With a pilot analysis on the initial $20$k responses, we find that the story generated in the second inference call is on average more diverse (refer to \Cref{tab:app:seq_prompt_increase_div}). These results motivated us to set up the final two-step data generation process, first inference call to collect natural responses from the model, and second inference call to redraft the natural response into a more diverse story. In the final data generation phase, we used $20$k unique three-word sets to generate prompts and sampled 10 first and second responses for each prompt, resulting in a dataset of 200,000 tuples of prompt, first response, and second response, per model (Llama-8B and Olmo-7B). We denote the data as follows: $\mathcal{D}^{(\pi)} = \{ (p, r_{1}, r_{2})_{i} \mid i = 1, \dots, 200{,}000 \}$
\noindent 
where, $p$, $r_1$, and $r_2$ denote the prompt, first response, and second response, respectively, generated from model (policy) $\pi$. Note that $|\{ p_1, p_2, \dots, p_{200,000} \}| = 20,000$ and we use two models, $m \in \{\text{Llama-8B, Olmo-7B} \}$, for data generation. 

\subsection{Data Filtration}
\label{sec:data_filter}

\paragraph{The Chosen and Rejected Pools.} 
Each instance in our generated dataset is a tuple \((p, r_1, r_2)\), where \(p\) is the prompt and \(r_1, r_2\) are two responses conditioned on it. The first response \(r_1\) reflects the model’s default behavior which are stories generated without intervention, capturing its most likely completion. In contrast, the second response \(r_2\) is generated with additional instructions aimed at reducing repetition, resulting in a more diverse output. We leverage this contrast by designating \(r_1\) as the \emph{rejected} response and \(r_2\) as the \emph{chosen} one. This setup encourages the model to prefer more diverse continuations that it is already capable of generating. Hence, it provides a strong self-learning framework for improving diversity.

\paragraph{Filtration Rules.}
Each pair \((r_1, r_2)\) gives us a natural candidate for rejected and chosen responses. On average, the second response $r_2$ is more diverse than the first $r_1$ (\Cref{tab:app:seq_prompt_increase_div}), but not every pair guarantees learning higher diversity. To ensure that the model receives consistent and useful learning signals, we apply a set of filtering rules.

First, we require that the diversity of $r_2$ exceeds that of $r_1$, so that the model consistently learns to prefer more diverse continuations. However, higher diversity may negatively impact text quality as prior work has shown a trade-off between the two \citep{zhang2020trading}. To ensure that preference learning also promotes higher quality, we further require that $r_2$ be of higher quality than $r_1$. Additionally, we filter out cases where both $r_1$ and $r_2$ are of poor quality, even if $r_2$ is marginally better. To do so, we enforce that $r_2$ must surpass the median quality of all $r_1$ responses. Lastly, most diversity metrics have been shown to be negatively correlated with text length \citep{covington2010cutting, shaib2024standardizing, mccarthy2010mtld}, which introduces a bias toward shorter texts. This issue has not been explicitly addressed in the recent studies for training and evaluation of LMs for diversity \citep{qin2025dive, lanchantin2025diverse, chung2025modifying}. To control for this, we constrain $r_1$ and $r_2$ to be of comparable length ($\pm5$ words). Ideally, we would like \(r_1\) and \(r_2\) to have exactly the same length. However, in practice, very few examples satisfy this strict constraint, especially when working with smaller language models (under 10B parameters). 
Therefore, we relax the constraint and allow a maximum length difference of \(\pm 5\) words between \(r_1\) and \(r_2\).

\begin{table*}[t]
\centering
\small
\setlength{\tabcolsep}{4pt}

\begin{tabular}{l|c|ccccc}
\toprule
\textbf{Method} & \textbf{Word Count} & \textbf{TTR} & \textbf{MATTR} & \textbf{HD-D} & \textbf{MTLD} & \textbf{MAAS} \\
\midrule
Entropy & $-0.1574$ & $\mathbf{0.2027}$ & $0.0800$ & $0.1071$ & $0.0656$ & $-0.1104$ \\
ArmoRM Score & $-0.3461$ & $0.1698$ & $-0.0042^{**}$ & $-0.0487$ & $0.0749$ & $\mathbf{0.2357}$ \\
\bottomrule
\end{tabular}


\caption{\small \textbf{Correlation Analysis.} Pearson correlation coefficients between six text statistics and two target metrics: entropy (diversity) and ArmoRM reward scores (quality). Both entropy and ArmoRM scores show negative correlation with text length. Among diversity metrics, TTR exhibits the strongest correlation with entropy, while the MAAS index shows the highest correlation with ArmoRM scores. $^{**}$: \(p < 0.001\); all others: \(p < 0.0001\).}
\label{tab:correlation_analysis}
\end{table*}

In summary, we retain a data point for preference learning only if it satisfies all of the following conditions, applied in order:
\begin{itemize}[itemsep=0pt, topsep=0pt, partopsep=0pt, leftmargin=*]
\label{filter_rules}
  \item The quality of $r_2$ is greater than or equal to the $50^{\text{th}}$ percentile of all $r_1$ quality scores.
  \item The quality of $r_2$ is greater than $r_1$.
  \item The diversity of $r_2$ is greater than $r_1$.
  \item The absolute difference in word count between $r_1$ and $r_2$ is at most five words.
\end{itemize}

\paragraph{Diversity and Quality Metrics.}
We use entropy to measure diversity and the ArmoRM reward model scores \citep{wang2024interpretable} to assess quality\footnote{refer to \Cref{sec:app:armo_rm} for a brief description of ArmoRM scores}. Entropy is a standard metric for lexical diversity \citep{lanchantin2025diverse}, with higher values indicating greater diversity. In our self-learning setup, entropy is useful because it reflects the model’s likelihood of producing a certain continuation of the prompt. When used in filtering, it helps identify training data that aligns with the model's own capabilities. For each example, we compute the entropy and the reward model score of both $r_1$ and $r_2$, conditioned on the original prompt $p$. 
When we use our data generation method, and use entropy and ArmoRM values for filtration, we call our approach,  Diverse, not Short (Diverse-NS or D-NS).

\begin{table}[h]
\centering

{
\small

\begin{tabular}{lcc}
\toprule
\textbf{} & \textbf{Num. Pref.} & \textbf{Word Count} \\
\textbf{Method} & \textbf{Pairs} &  $\Delta$ \\
\midrule
No Filtering        & 200{,}000 & $-0.68 \pm \text{\scriptsize 11.33}$ \\
DivPO               & 3{,}000 & $-49.90 \pm \text{\scriptsize 17.51}$ \\
Ours - D-NS-Lite    & 3{,}000 & $-0.90 \pm \text{\scriptsize 2.91}$ \\
Ours - D-NS         & 3{,}000 & $-1.35 \pm \text{\scriptsize 2.93}$ \\
\bottomrule
\end{tabular}

}


\caption{\small \textbf{Data Properties After Filtering.}  
This table reports the average ($\pm$std.dev.) length difference (\(\Delta\)) between \textit{chosen} and \textit{rejected}. While DivPO tends to favor significantly shorter \textit{chosen} responses.}
\label{tab:post_filter_stats}
\end{table}

\paragraph{Lightweight Filtration.}  
While entropy and ArmoRM scores are high-quality metrics for measuring diversity and response quality, they are computationally expensive. Each example \((p, r_1, r_2)\) requires two additional inference calls to compute entropy and two more for ArmoRM scoring. 
To reduce this overhead, we evaluated seven alternative metrics and measured their correlation with entropy and ArmoRM scores. Among these, Type-Token Ratio (TTR) showed the highest correlation with entropy (Pearson \(r = 0.2027\), \(p < 0.0001\)), and the MAAS index \citep{mass1972zusammenhang} was most correlated with ArmoRM scores (Pearson \(r = 0.2357\), \(p < 0.0001\)). Refer to \Cref{tab:correlation_analysis} for all correlation results. Based on these findings, we replace entropy with TTR and ArmoRM scores with MAAS in our filtering pipeline. When this lightweight variant is used during data filtering, we refer to the resulting method as Diverse-NS-Lite (or D-NS-Lite). We provide a discussion on the computational savings achieved with Diverse-NS-Lite in \Cref{sec:app:computational_saving_d_ns_lite}.

\paragraph{Post-Filtration Properties.}
Based on the correlation analysis (\Cref{tab:correlation_analysis}), it is worth noting that both entropy and ArmoRM scores are negatively correlated with text length. As a result, optimizing for diversity or quality alone may unintentionally favor shorter responses as the ``chosen'' continuations. To avoid this bias, it is essential to explicitly control for length when curating preference learning data for improving diversity. 
To show this, we implement a recent study that is closest to our method, Diverse Preference Optimization(DivPO) \citep{lanchantin2025diverse} (refer to \Cref{sec:app:divpo_implementation} for implementation details). DivPO also generates responses and filters the responses to form preference learning pairs without explicitly controlling the length of the chosen and rejected continuations. We compare pre- and post-filtration data properties for DivPO and Diverse-NS in Tab. \ref{tab:post_filter_stats}. The table clearly shows that in the pursuit of maximizing the entropy values, DivPO selects significantly shorter (-$49.90$ words shorter on average) responses as the \textit{chosen} responses in the final preference data. 

\section{Experimental and Evaluation Setup}

\subsection{Preference Tuning}

After generating and filtering the data, we fine-tune the same base policy \(\pi\) that was used to generate it. In other words, data generated by Llama-8B is used to train Llama-8B, and likewise for Olmo-7B. To ensure a fair comparison across methods (DivPO, D-NS, and D-NS-Lite), we limit the final training dataset to 3,000 preference pairs\footnote{We observed that the size of the dataset after filtering is the smallest for Diverse-NS, slightly more than $3$k. Hence, to make the training runs more comparable across methods, we limit the size of the dataset to $3$k for all methods.}. To construct this $3$k dataset, we first compute the entropy gain for each pair as the difference between the entropy of the \textit{chosen} and \textit{rejected} responses \footnote{note that, by construction, the \textit{chosen} response has higher entropy in the filtered set}. We then sort all pairs by entropy gain in descending order and select the top $3$k examples. This ensures that the final training set maximizes diversity gain for the base model. The same selection procedure is applied to all three methods. 

We further extend our experiments to evaluate the utility of training larger models with data generated from smaller ones. For this, we train Olmo-13B using preference pairs generated from Olmo-7B. We provide all hyperparameter values in \Cref{sec:app:hyperparameters}.
All experiments are run on a single NVIDIA RTX 6000 GPU (48GB memory), using a per-device batch size of 2 and a global batch size of 64. Training Llama-8B or Olmo-7B takes approximately 100–150 minutes while O-13B takes 200-220 minutes per run, highlighting our setup efficiency.

\subsection{Evaluation}

\subsubsection{Tasks}
We evaluate the model’s response diversity with four tasks: Divergent Association Task (DAT), Persona Generation Task (PGT), Alternate Uses Task (AUT), and Creative Writing Task (CWT).

\paragraph{Divergent Associations Task (DAT).} 
The DAT \citep{olson_naming_2021} is a psychological test commonly used to assess divergent thinking in humans. Participants are asked to generate a list of 10 words that are as dissimilar from each other as possible. Recent studies have adapted DAT to evaluate the creativity of language models, focusing on their ability to produce diverse outputs \citep{bellemare2024divergent}. To quantify model performance on DAT, we use the Divergent Semantic Integration (DSI) metric \citep{johnson2023divergent}, which computes the average semantic distance of each word in the generated list from all others. Higher DSI values indicate more divergent thinking and greater ideological diversity. Following \citet{johnson2023divergent}, we extract token embeddings from the $6^{th}$ 
layer of BERT-large for the generated list and compute the average pairwise cosine distance between all embeddings. This approach has been shown to correlate strongly with human judgments of creativity \citep{johnson2023divergent}. We provide the exact prompt used for DAT in \Cref{sec:app:prompts_model_eval}.
For a robust evaluation, we sample 100 DAT responses per model using temperature 1.0 and different random seeds. From these 100 lists (each with 10 words), we compute and report two metrics: (1) the average and standard deviation of DSI scores, and (2) the number of unique words across all 1,000 generated tokens. In both cases, higher values indicate greater diversity.

\paragraph{Persona Generation Task (PGT).}
To assess diversity in structured generation, we use the PGT, also used in the study conducted by \citet{lanchantin2025diverse}. In this task, the model is prompted to generate a JSON object with three fields: first name, city of birth, and current occupation to evaluate the model’s ability to produce varied persona descriptions. The exact prompt is provided in \Cref{sec:app:prompts_model_eval}. We sample 100 responses per model using temperature 1.0 and different random seeds. For each key in the JSON object, we report the proportion of unique values across the 100 responses. Higher uniqueness indicates greater diversity.

\paragraph{Alternate Uses Task (AUT).}
The Alternate Uses Task (AUT) is a common and rigorously validated psychological test to measure human divergent thinking \citep{guilford_structure_1956}. In this task, the subject/model is asked to generate creative and unconventional uses for objects (e.g., broom). The prompt and list of objects used for evaluation are provided in \Cref{sec:app:prompts_model_eval}. We use 15 unique objects and generate 10 responses per object using different random seeds, resulting in 150 total responses sampled at temperature 1.0. For quantifying the diversity of the generated uses, we measure the distance between the target object and generated uses with the help of BERT-large encodings, a validated approach that correlates with human creativity ratings \citep{patterson2023multilingual}. We report the mean and standard deviation of the distance values, higher values indicate higher diversity. 

\paragraph{Creative Writing Task (CWT).}
The CWT — based on a well-validated psychological assessment of creativity \citep{prabhakaran_thin_2014} — is exactly the same as our data generation task. That is, given a set of three words, the subject/model is tasked with generating a creative short story that includes all three words. We provide a separate list of three-word sets used for evaluation in \Cref{sec:app:prompts_model_eval}. We sample 10 responses for each of the seven three-word sets with temperature of $1.0$. Unlike our other evaluation tasks, we measure the diversity as well as the quality of the generated responses. Similar to \citet{johnson2023divergent}, we calculate the DSI metric to measure the diversity of the generated story. For quality measurements, we resort to the ArmoRM reward model preference scores \citep{wang2024interpretable}. 
We report DSI, ArmoRM, and 4-gram diversity values, where higher values are more desirable for all metrics. 

\begin{table*}[t]
\centering
\small
\begin{tabular}{llcc|cc}
\toprule

&&&& \multicolumn{2}{c}{\textbf{Ours}} \\

\textbf{Task} & \textbf{Metric} & \textbf{Base Model} & \textbf{DivPO} & \textbf{D-NS-Lite} & \textbf{D-NS} \\
\midrule

\multicolumn{6}{c}{\textbf{LLaMA-8B}} \\

\midrule
\midrule

DAT & DSI 
    & $0.7535$ 
    & $0.7545$ 
    & $0.7590$ 
    & $\mathbf{0.7640}$ \\
DAT & Unique Words 
    & $0.4575$ & $0.4593$ & $0.4797$ & $\mathbf{0.4914}$ \\

PGT & Unique First Names 
    & $0.6500$ & $0.6100$ & $\mathbf{0.6900}$ & $\mathbf{0.6900}$ \\
PGT & Unique Cities 
    & $0.3300$ & $0.3100$ & $\mathbf{0.4700}$ & $0.4200$ \\
PGT & Unique Occupations 
    & $0.4100$ & $0.3900$ & $\mathbf{0.5100}$ & $0.4900$ \\

AUT & DSI 
    & $0.8876$ 
    & $0.8837$ 
    & $0.8876$ 
    & $\mathbf{0.8878}$ \\

CWT & DSI 
    & $0.8515$ 
    & $0.8521$ 
    & $0.8556$ 
    & $\mathbf{0.8581}$ \\
CWT & ArmoRM Score 
    & $0.1451$ 
    & $\mathbf{0.1495}$ 
    & $0.1369$ 
    & $0.1405$ \\
CWT & 4-gram div.
    & 2.8550
    & 2.9320
    & 2.9450
    & \textbf{2.9620} \\

\midrule
\multicolumn{6}{c}{\textbf{OLMo-7B}} \\
\midrule
\midrule

DAT & DSI 
    & $0.7480$ 
    & $0.7509$ 
    & $\mathbf{0.7662}$ 
    & $0.7639$ \\
DAT & Unique Words 
    & $0.6139$ & $0.6079$ & $\mathbf{0.6347}$ & $0.6327$ \\

PGT & Unique First Names 
    & $0.3300$ & $0.3300$ & $0.3300$ & $\mathbf{0.3400}$ \\
PGT & Unique Cities 
    & $\mathbf{0.3100}$ & $0.3000$ & $0.2700$ & $0.2700$ \\
PGT 
    & Unique Occupations & $0.5200$ & $0.5500$ & $\mathbf{0.6100}$ & $\mathbf{0.6100}$ \\

AUT & DSI 
    & $0.8836$ 
    & $0.8846$ 
    & $0.8852$ 
    & $\mathbf{0.8858}$ \\

CWT & DSI 
    & $0.8499$ 
    & $0.8491$ 
    & $0.8548$ 
    & $\mathbf{0.8563}$ \\
CWT & ArmoRM Score 
    & $0.1435$ 
    & $0.1441$ 
    & $0.1462$ 
    & $\mathbf{0.1464}$ \\
CWT & 4-gram div.
    & 3.1270
    & 3.1690
    & \textbf{3.1750}
    & 3.1620 \\

\midrule
\multicolumn{6}{c}{\textbf{OLMo-13B}} \\
\midrule
\midrule

DAT & DSI 
    & $0.7233$ 
    & $0.7282$ 
    & $0.7320$ 
    & $\mathbf{0.7364}$ \\
DAT & Unique Words 
    & $\mathbf{0.3421}$ & $0.3340$ & $0.3310$ & $0.3256$ \\
    
PGT & Unique First Names 
    & $0.4100$ & $0.4100$ & $0.4400$ & $\mathbf{0.4500}$ \\
PGT & Unique Cities 
    & $0.3500$ & $0.3500$ & $0.3700$ & $\mathbf{0.3900}$ \\
PGT & Unique Occupations 
    & $0.1900$ & $0.1900$ & $0.1900$ & $\mathbf{0.2000}$ \\

AUT & DSI 
    & $0.8943$ 
    & $0.8960$ 
    & $\mathbf{0.8974}$ 
    & $0.8970$ \\

CWT & DSI 
    & $0.8557$ 
    & $0.8555$ 
    & $\mathbf{0.8616}$ 
    & $0.8614$ \\
CWT & ArmoRM Score 
    & $0.1571$ 
    & $0.1589$ 
    & $0.1585$ 
    & $\mathbf{0.1590}$ \\
CWT & 4-gram div.
    & 3.0820
    & 3.0770
    & 3.095
    & \textbf{3.1070} \\

\bottomrule
\end{tabular}

\caption{\small \textbf{Diversity and Quality Evaluation.} We present the average diversity (DSI or unique values) and quality (ArmoRM Score) measurements for model responses collected on four creative generation tasks (Structured Gen.: DAT, PGT, Free-Form Gen.: AUT, CWT).}
\label{tab:all_combined_results}
\end{table*}

\subsubsection{Length-Adjusted Evaluation}
While most diversity metrics exhibit bias toward shorter outputs, \citet{johnson2023divergent} shows that the DSI metric displays the opposite tendency-it favors longer responses. This is not an issue in tasks like DAT, where the output length is fixed at 10 words. But for open-ended tasks such as CWT, longer stories may receive disproportionately high DSI scores primarily due to their length, rather than genuine diversity. To address this issue, we introduce a novel evaluation metric: \textit{$\Delta$ Diversity Decile ($\Delta$ DD)}, which accounts for text length when assessing diversity.

\paragraph{Change in Diversity Decile ($\Delta$DD).}
We first build a \emph{decile map} that captures the empirical distribution of diversity scores at each length.  Using 800\,000 stories collected from Llama-8B and Olmo-7B over 40\,000 prompts, we:
(1) group responses by word count $w$;
(2) compute decile thresholds for a chosen diversity metric (e.g.\ TTR, MTLD); and
(3) store these percentile thresholds in a lookup table $\mathcal{M}$.
Here, a \emph{decile} refers to one of ten intervals that divide the distribution of diversity scores for a given length into ten equal parts. The top decile corresponds to the most diverse 10\% of responses at that length, the second-highest to the next 10\%, and so on. This mapping allows us to estimate the \emph{approximate diversity rank} of any new response relative to other responses of the same length. At evaluation time, a new response $r$ with word count $w_r$ and diversity score $d_r$ is assigned the highest decile index $k\in\{0,\dots,9\}$ such that $d_r$ exceeds the $k$-th threshold in $\mathcal{M}[w_r]$.  Formally,
$DD(r,\mathcal{M}) \;=\; k,$
where larger $k$ means the response is more diverse than a greater share of previously observed texts of the \emph{same length}\footnote{If the response length of the new response is not present in the constructed map, we select the closest neighbor as an approximation. However, it is important to note that due to the ample size of our synthetic data, we never encounter a case where the response length value is not found in the constructed map.}. 

To evaluate the effect of preference tuning, we average DD scores over 70 CWT prompts for the base and the preference-tuned models and report their difference: $\Delta\!DD = \overline{DD}_{\text{tuned}} - \overline{DD}_{\text{base}}.$

Positive $\Delta DD$ values indicate improved diversity, with higher values corresponding to a larger improvement. Negative values signify reduced diversity, and $\Delta DD=0$ signifies no change.
Note that, DD is agnostic to the choice of diversity metric. We therefore report \(\Delta\)DD values using seven standard metrics: TTR, MATTR, HD-D, MTLD, and MAAS. We also compute \textit{\(\Delta\)DD} using ArmoRM reward scores to quantify the gain or loss in quality.
This length-aware normalization prevents either long or short responses from being over-credited for diversity\footnote{We provide a summary of all metrics in \Cref{tab:app:metrics}}.

\begin{figure*}
    \centering
    \includegraphics[width=0.8\linewidth]{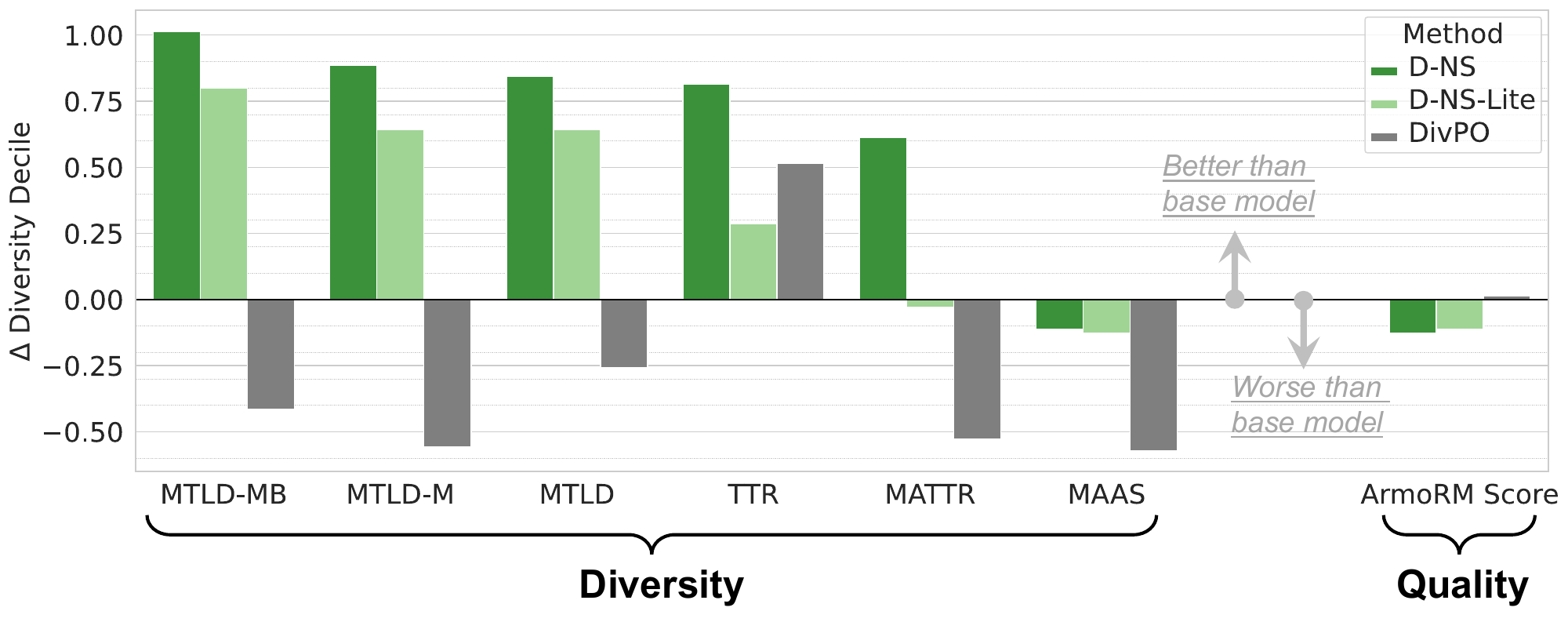}
    \caption{\small \textbf{Diversity and Quality Evaluation on CWT.}  
    This figure shows \(\Delta\)Diversity Decile (\(\Delta DD\)) values (y-axis) across various metrics (x-axis), computed from 70 CWT responses generated by the Olmo-2-7B model. A value of zero represents base model performance; bars indicate improvements from preference-tuned models. \textit{D-NS} achieves the highest diversity gains overall, while \textit{D-NS-Lite} consistently outperforms \textit{DivPO}, except under TTR. In terms of quality (ArmoRM), \textit{DivPO} shows a slight improvement, whereas our methods show a minor drop.}

    \label{fig:cwt:diversity_decile}
\end{figure*}

\section{Results}
\label{section:results}

\paragraph{Divergent Associations Task (DAT).}
In our DAT evaluation (Tab. \ref{tab:all_combined_results}), we see that both Diverse-NS and its lightweight variant deliver clear improvements in diversity over the untrained base and the DivPO baseline across all model sizes. Remarkably, even the D-NS-Lite variant consistently outperforms DivPO, demonstrating that a compact diversity strategy can be highly effective. Interestingly, using data generated by the smaller Olmo-7B to fine-tune the larger Olmo-13B yields diversity gains for every method, highlighting how smaller models can serve as powerful ``diversity teachers'' for their larger counterparts.

\paragraph{Persona Generation Task (PGT).}
In our PGT evaluation (Tab. \ref{tab:all_combined_results}), Diverse-NS produces more distinct first names, cities, and occupations than DivPO for every model, with the sole exception of the city metric on Olmo-7B. Outside that one case, Diverse-NS-Lite also outperforms DivPO across all three metrics. Notably, on Llama-8B, Diverse-NS-Lite matches or exceeds the baseline and Diverse-NS on every attribute of the task.

\paragraph{Alternate Uses Task (AUT). }
In our AUT evaluation (Tab. \ref{tab:all_combined_results}), Diverse-NS-Lite consistently beats DivPO, and Diverse-NS consistently beats Diverse-NS-Lite, though only by a small margin.

\paragraph{Creative Writing Task (CWT). }

In our CWT evaluations (Tab. \ref{tab:all_combined_results}), Diverse-NS produces the highest DSI scores for both Llama-8B and Olmo-7B. Interestingly, for Llama-8B the other methods actually reduce the ArmoRM score below baseline but Diverse-NS exceeds it. The highest 4-gram diversity is observed for Diverse-NS or -Lite in all cases.
We also compute 
$\Delta DD$ with six lexical diversity measures and ArmoRM. Both Diverse-NS and its lightweight variant significantly outperform DivPO on every diversity metric. 
The $\Delta DD$ remains above the baseline for all metrics except MAAS, where it dips marginally below and similarly shows a slight under-performance for ArmoRM. Crucially, even where $\Delta DD$ suggests a minor quality drop, the absolute diversity values after self-training still exceed those of the base model (despite longer outputs), indicating that any loss in writing quality is minimal (refer to \Cref{sec:app:dd_eval} for Llama-8B and Olmo-13B results)\footnote{We provide all results with std. dev. values in \Cref{tab:app:all_combined_results}}.

\paragraph{Significance of DSI Improvements.} 
In line with the findings of \citet{johnson2023divergent}, 
we observe a narrow distribution of DSI values in our CWT experiments (percentiles: $5^{th},95^{th}: 0.84, 0.87$).
Given the narrow distribution, to highlight the significance of the DSI improvements achieved with D-NS or D-NS-Lite, we conduct an independent t-test on method pairs, e.g., base model versus D-NS, with 70 DSI measurements (for 70 unique CWT responses) for each method. 
Results are presented in \Cref{tab:ttest-cwt}, where negative t-statistics indicate that ``Method B'' yields higher response diversity than ``Method A''. We find that both D-NS and D-NS-Lite produce significantly more diverse responses than the base model and DivPO. On the other hand, the differences between the base model and DivPO, as well as between D-NS and D-NS-Lite, are not statistically significant. These findings further underscore the effectiveness of D-NS-Lite as a computationally efficient alternative to D-NS.

\begin{table}[t]
\centering
\small
\begin{tabular}{@{}llcc@{}}
\toprule
\textbf{Method A} & \textbf{Method B} & \textbf{t-statistic} & \textbf{p-value} \\
\midrule
\multicolumn{4}{c}{\textbf{LLaMA-8B}} \\
\midrule
\midrule
Base    & DivPO      & $-0.4126$  & Not Significant \\
Base    & D-NS       & \cellcolor{lightgreen} $-4.1208$  & $< 0.001$       \\
Base    & D-NS-Lite  & \cellcolor{lightgreen} $-2.6100$  & $< 0.05$        \\
DivPO   & D-NS       & \cellcolor{lightgreen} $-3.7516$  & $< 0.001$       \\
DivPO   & D-NS-Lite  & \cellcolor{lightgreen} $-2.2214$  & $< 0.05$        \\
D-NS    & D-NS-Lite  & 1.5492   & Not Significant \\
\midrule
\multicolumn{4}{c}{\textbf{Olmo-7B}} \\
\midrule
\midrule
Base    & DivPO      & $0.3609$   & Not Significant \\
Base    & D-NS       & \cellcolor{lightgreen} $-3.4402$  & $< 0.001$       \\
Base    & D-NS-Lite  & \cellcolor{lightgreen} $-2.7348$  & $< 0.01$        \\
DivPO   & D-NS       & \cellcolor{lightgreen} $-3.5529$  & $< 0.001$       \\
DivPO   & D-NS-Lite  & \cellcolor{lightgreen} $-2.8998$  & $< 0.01$        \\
D-NS    & D-NS-Lite  & $0.9972$   & Not Significant \\
\bottomrule
\end{tabular}
\caption{\small \textbf{Significance of DSI Improvements.} In this table, we present the results of independent t-tests on DSI scores between pairs of methods. Negative t-values indicate that ``Method B'' yields more diverse responses than ``Method A''. We highlight cases (in green) where our proposed methods (D-NS or D-NS-Lite) are significantly better than the base model or DivPO.}
\label{tab:ttest-cwt}
\end{table}

\paragraph{Human Evaluation.} 
To further understand the human perceived value of DSI improvements, we conduct a human evaluation on the CWT responses. Under pairwise evaluation, annotators are shown a prompt along with two responses from two different methods (e.g., D-NS versus DivPO) and asked to select the more diverse response (see \Cref{sec:app:prompts_model_eval}). 
Per CWT prompt, we select 20 least similar (Jaccard sim.) pairs to ensure maximally distinguishable comparisons between methods.
Using 7 prompts, we construct 140 pairs each for D-NS vs. Base and D-NS vs. DivPO. The resulting 280 comparisons are evenly and randomly assigned to four annotators: two involved in the study (E1, E2) and two independent evaluators with no prior exposure (E3, E4).
Human evaluation confirms that D-NS consistently outperforms both baselines in win percentage. Except for E3, all annotators strongly preferred D-NS, with win rates reaching as high as $72.73\%$ against DivPO (E4). These findings suggest that even modest improvements in DSI scores correspond to clear and consistent gains in human-perceived diversity. We extended our human evaluation to the LLM-as-a-Judge evaluation that covers comparison between D-NS-Lite and the baselines as well. The results of the LLM-as-a-Judge experiment reciprocate the human evaluation findings and underscore the effectiveness of both D-NS and D-NS-Lite (refer to \Cref{sec:app:llm_as_a_judge} for details). We provide a few examples of the model responses in \Cref{sec:app:cwt_response_examples} for the readers.

\section{Discussion}
\label{sec:discussion}

We introduced \emph{Diverse-NS}, a 
data selection strategy
to improve output diversity while preserving quality. Experiments with Llama-8B and Olmo-7B show that \emph{Diverse-NS} improves diversity on four creative generation tasks: DAT, PGT, AUT, CWT. 
On CWT, the diversity gains achieved by \emph{Diverse-NS} and its lightweight variant, \emph{Diverse-NS-Lite}, are statistically significant. These improvements are further supported by human evaluations, where \emph{Diverse-NS} achieves higher win rates against both the base model and DivPO.

\begin{table}[t]
\centering
\small
\begin{tabular}{@{}lcc@{}}
\toprule
\textbf{Human} & \textbf{D-NS vs. Base} & \textbf{D-NS vs. DivPO} \\
\textbf{Evaluator} & (D-NS Win\% - Tie\%) & (D-NS Win\% - Tie\%) \\
\midrule
E1 & $63.16 - 28.95$ & $69.70 - 24.24$ \\
E2 & $72.22 - \ 0.00$  & $63.64 - \ 0.00$  \\
E3 & $50.00 - \ 5.88$  & $51.52 - \ 3.03$  \\
E4 & $64.71 - 11.76$ & $72.73 - 15.15$ \\
\bottomrule
\end{tabular}
\caption{\small \textbf{Human Evaluation on CWT.} Pairwise human evaluation results comparing D-NS against the base model and DivPO on the CWT task. Each cell shows the win and tie percentages for D-NS. D-NS consistently achieves higher win rates against both baselines.}
\label{tab:human_eval_cwt}
\end{table}

\paragraph{\emph{Diverse-NS} is highly efficient.}
All gains are achieved with only $3$k preference pairs and less than two hours of training on a single 48 GB GPU. The lightweight variant, \emph{Diverse-NS-Lite}, replaces costly entropy and ArmoRM scoring with inexpensive proxies yet still surpasses DivPO in nearly every setting. We further show that a 7B model can act as an effective ``diversity teacher'' for its 13B counterpart, pointing to a low-cost path for diversity-aware alignment at scale.

\paragraph{\emph{Diverse-NS} maintains high quality.} Diversity and quality are often at odds \citep{zhang2020trading, chung2023increasing}, and we observe this trade-off in our experiments as well. However, there are encouraging instances where both improve together. For Olmo-7B and Olmo-13B, the ArmoRM score increases alongside diversity. $\Delta$Diversity Decile values further confirms that, for Olmo-13B, diversity and quality consistently rise in tandem. In other cases, we observe only a minor drop in quality, suggesting that \emph{Diverse-NS} effectively balances this trade-off in most scenarios.  


\paragraph{The long-standing challenge of length.} Evaluating diversity remains difficult due to the well-known length bias in most diversity metrics. This issue extends to ArmoRM scores, which also favor shorter texts (Tab. \ref{tab:correlation_analysis}), further complicating reliable evaluation. To mitigate this, we introduce the \emph{$\Delta$Diversity Decile} metric, which quantifies percentile gains or losses in diversity (or quality) relative to the base model. Using this length-adjusted metric, we observe substantial improvements in diversity across most lexical diversity measures, along with small but mixed changes in quality.

Overall, \emph{Diverse-NS} offers a practical and scalable solution for boosting diversity in aligned LLMs. By addressing the length bias in both training and evaluation, our  sets a foundation for more expressive and diverse language generation. We hope this work encourages further exploration of length-aware diversity alignment.


\section*{Limitations}
While our study demonstrates the effectiveness of diversity-aware self-learning, several areas remain open for future exploration. First, our data filtering relies on a single diversity metric (e.g., entropy or TTR). Although effective, no single metric can fully capture all aspects of text diversity. Future work could incorporate multiple metrics to jointly optimize lexical, semantic, and syntactic variation, as well as novelty, to better capture diverse training signals.
Second, we focus on one data generation task—short story writing—which allows for controlled analysis and task-specific improvements. Expanding the framework to include a broader set of tasks could lead to more generalizable diversity enhancements.
Third, our self-learning setup investigates only a single round of preference tuning. While this provides a strong baseline, recent work suggests that repeated rounds of self-training can affect diversity \citep{guo2023curious, seddik2024bad, herel2024collapse}. It would be valuable to study how diversity evolves across multiple self-learning iterations in our framework.
It is worth noting a peculiar change in the length distribution of the preference-tuning model (\Cref{tab:app:post_training_response_length}). Even though preference pairs are of comparable lengths in \emph{Diverse-NS} and \emph{Diverse-NS-Lite}, the model learns to be more expressive. We suspect this shift is influenced by a skewed proportion of longer preference pairs, which may inadvertently bias the model toward generating longer responses. Controlling the length distribution is challenging under our current framework due to the strict filtering criteria. In future work, we aim to address this by extending our method to a multi-task setup that includes both short and long generation tasks.

\section*{Ethics Statement}
Our work focuses on improving the diversity of language model outputs, particularly in creative and open-ended tasks. While diversity is an important dimension of language generation, it may come at the cost of factual correctness in certain scenarios. Therefore, we caution against the use of our dataset or models in tasks where factual accuracy is critical, such as medical advice, legal reasoning, or scientific fact-checking.
We also acknowledge the growing computational divide in language model research. A key motivation behind our approach is to make diversity-aware alignment more accessible. By limiting training to 3,000 preference pairs and demonstrating the effectiveness of smaller models (e.g., Olmo-2-7B) as diversity teachers, we aim to lower the resource barrier and encourage further research in compute-constrained environments.
Finally, while we use proprietary language models (such as GPT-4o and Claude) to assist in editing and refining text during data curation and paper writing, no portion of this manuscript was generated entirely by an LLM. All content has been written, reviewed, and edited by the authors to ensure clarity, originality, and scientific rigor.

\section*{Acknowledgment}
This work was funded in part by an Amazon AGI research award to Anna Rumshisky.

\bibliography{references}

@article{west2025base,
  title={Base Models Beat Aligned Models at Randomness and Creativity},
  author={West, Peter and Potts, Christopher},
  journal={arXiv preprint arXiv:2505.00047},
  year={2025}
}

@article{kirk2023understanding,
  title={Understanding the effects of rlhf on llm generalisation and diversity},
  author={Kirk, Robert and Mediratta, Ishita and Nalmpantis, Christoforos and Luketina, Jelena and Hambro, Eric and Grefenstette, Edward and Raileanu, Roberta},
  journal={arXiv preprint arXiv:2310.06452},
  year={2023}
}

@article{doshi2024generative,
  title={Generative AI enhances individual creativity but reduces the collective diversity of novel content},
  author={Doshi, Anil R and Hauser, Oliver P},
  journal={Science Advances},
  volume={10},
  number={28},
  pages={eadn5290},
  year={2024},
  publisher={American Association for the Advancement of Science}
}

@article{padmakumar2023does,
  title={Does writing with language models reduce content diversity?},
  author={Padmakumar, Vishakh and He, He},
  journal={arXiv preprint arXiv:2309.05196},
  year={2023}
}

@inproceedings{anderson2024homogenization,
  title={Homogenization effects of large language models on human creative ideation},
  author={Anderson, Barrett R and Shah, Jash Hemant and Kreminski, Max},
  booktitle={Proceedings of the 16th conference on creativity \& cognition},
  pages={413--425},
  year={2024}
}

@article{shaib2024detection,
  title={Detection and measurement of syntactic templates in generated text},
  author={Shaib, Chantal and Elazar, Yanai and Li, Junyi Jessy and Wallace, Byron C},
  journal={arXiv preprint arXiv:2407.00211},
  year={2024}
}

@article{xu2024echoes,
  title={Echoes in AI: Quantifying Lack of Plot Diversity in LLM Outputs},
  author={Xu, Weijia and Jojic, Nebojsa and Rao, Sudha and Brockett, Chris and Dolan, Bill},
  journal={arXiv preprint arXiv:2501.00273},
  year={2024}
}

@article{shumailov2023curse,
  title={The curse of recursion: Training on generated data makes models forget},
  author={Shumailov, Ilia and Shumaylov, Zakhar and Zhao, Yiren and Gal, Yarin and Papernot, Nicolas and Anderson, Ross},
  journal={arXiv preprint arXiv:2305.17493},
  year={2023}
}

@article{guo2023curious,
  title={The curious decline of linguistic diversity: Training language models on synthetic text},
  author={Guo, Yanzhu and Shang, Guokan and Vazirgiannis, Michalis and Clavel, Chlo{\'e}},
  journal={arXiv preprint arXiv:2311.09807},
  year={2023}
}

@article{seddik2024bad,
  title={How bad is training on synthetic data? a statistical analysis of language model collapse},
  author={Seddik, Mohamed El Amine and Chen, Suei-Wen and Hayou, Soufiane and Youssef, Pierre and Debbah, Merouane},
  journal={arXiv preprint arXiv:2404.05090},
  year={2024}
}

@article{herel2024collapse,
  title={Collapse of Self-trained Language Models},
  author={Herel, David and Mikolov, Tomas},
  journal={arXiv preprint arXiv:2404.02305},
  year={2024}
}

@article{ouyang2022training,
  title={Training language models to follow instructions with human feedback},
  author={Ouyang, Long and Wu, Jeffrey and Jiang, Xu and Almeida, Diogo and Wainwright, Carroll and Mishkin, Pamela and Zhang, Chong and Agarwal, Sandhini and Slama, Katarina and Ray, Alex and others},
  journal={Advances in neural information processing systems},
  volume={35},
  pages={27730--27744},
  year={2022}
}

@inproceedings{longpre2023flan,
  title={The flan collection: Designing data and methods for effective instruction tuning},
  author={Longpre, Shayne and Hou, Le and Vu, Tu and Webson, Albert and Chung, Hyung Won and Tay, Yi and Zhou, Denny and Le, Quoc V and Zoph, Barret and Wei, Jason and others},
  booktitle={International Conference on Machine Learning},
  pages={22631--22648},
  year={2023},
  organization={PMLR}
}

@article{bai2022training,
  title={Training a helpful and harmless assistant with reinforcement learning from human feedback},
  author={Bai, Yuntao and Jones, Andy and Ndousse, Kamal and Askell, Amanda and Chen, Anna and DasSarma, Nova and Drain, Dawn and Fort, Stanislav and Ganguli, Deep and Henighan, Tom and others},
  journal={arXiv preprint arXiv:2204.05862},
  year={2022}
}

@article{dai2023safe,
  title={Safe rlhf: Safe reinforcement learning from human feedback},
  author={Dai, Josef and Pan, Xuehai and Sun, Ruiyang and Ji, Jiaming and Xu, Xinbo and Liu, Mickel and Wang, Yizhou and Yang, Yaodong},
  journal={arXiv preprint arXiv:2310.12773},
  year={2023}
}

@article{schulman2017proximal,
  title={Proximal policy optimization algorithms},
  author={Schulman, John and Wolski, Filip and Dhariwal, Prafulla and Radford, Alec and Klimov, Oleg},
  journal={arXiv preprint arXiv:1707.06347},
  year={2017}
}

@article{peng2023instruction,
  title={Instruction tuning with gpt-4},
  author={Peng, Baolin and Li, Chunyuan and He, Pengcheng and Galley, Michel and Gao, Jianfeng},
  journal={arXiv preprint arXiv:2304.03277},
  year={2023}
}

@article{olmo20242,
  title={2 OLMo 2 Furious},
  author={OLMo, Team and Walsh, Pete and Soldaini, Luca and Groeneveld, Dirk and Lo, Kyle and Arora, Shane and Bhagia, Akshita and Gu, Yuling and Huang, Shengyi and Jordan, Matt and others},
  journal={arXiv preprint arXiv:2501.00656},
  year={2024}
}

@article{grattafiori2024llama,
  title={The llama 3 herd of models},
  author={Grattafiori, Aaron and Dubey, Abhimanyu and Jauhri, Abhinav and Pandey, Abhinav and Kadian, Abhishek and Al-Dahle, Ahmad and Letman, Aiesha and Mathur, Akhil and Schelten, Alan and Vaughan, Alex and others},
  journal={arXiv preprint arXiv:2407.21783},
  year={2024}
}

@article{wang2024interpretable,
  title={Interpretable preferences via multi-objective reward modeling and mixture-of-experts},
  author={Wang, Haoxiang and Xiong, Wei and Xie, Tengyang and Zhao, Han and Zhang, Tong},
  journal={arXiv preprint arXiv:2406.12845},
  year={2024}
}

@misc{cosmopedia2024,
  author       = {Hugging Face},
  title        = {Cosmopedia: An open source mixture of experts for retrieval-augmented generation},
  year         = {2024},
  howpublished = {\url{https://huggingface.co/blog/cosmopedia}},
  note         = {Accessed: 2025-05-16}
}

@article{li2023textbooks,
  title={Textbooks are all you need ii: phi-1.5 technical report},
  author={Li, Yuanzhi and Bubeck, S{\'e}bastien and Eldan, Ronen and Del Giorno, Allie and Gunasekar, Suriya and Lee, Yin Tat},
  journal={arXiv preprint arXiv:2309.05463},
  year={2023}
}

@article{chen2024diversity,
  title={On the Diversity of Synthetic Data and its Impact on Training Large Language Models},
  author={Chen, Hao and Waheed, Abdul and Li, Xiang and Wang, Yidong and Wang, Jindong and Raj, Bhiksha and Abdin, Marah I},
  journal={arXiv preprint arXiv:2410.15226},
  year={2024}
}

@article{chung2023increasing,
  title={Increasing diversity while maintaining accuracy: Text data generation with large language models and human interventions},
  author={Chung, John Joon Young and Kamar, Ece and Amershi, Saleema},
  journal={arXiv preprint arXiv:2306.04140},
  year={2023}
}

@article{lu2024benchmarking,
  title={Benchmarking language model creativity: A case study on code generation},
  author={Lu, Yining and Wang, Dixuan and Li, Tianjian and Jiang, Dongwei and Khudanpur, Sanjeev and Jiang, Meng and Khashabi, Daniel},
  journal={arXiv preprint arXiv:2407.09007},
  year={2024}
}

@article{zhang2020trading,
  title={Trading off diversity and quality in natural language generation},
  author={Zhang, Hugh and Duckworth, Daniel and Ippolito, Daphne and Neelakantan, Arvind},
  journal={arXiv preprint arXiv:2004.10450},
  year={2020}
}

@article{tian2023macgyver,
  title={MacGyver: Are Large Language Models Creative Problem Solvers?},
  author={Tian, Yufei and Ravichander, Abhilasha and Qin, Lianhui and Bras, Ronan Le and Marjieh, Raja and Peng, Nanyun and Choi, Yejin and Griffiths, Thomas L and Brahman, Faeze},
  journal={arXiv preprint arXiv:2311.09682},
  year={2023}
}

@article{ge2024scaling,
  title={Scaling synthetic data creation with 1,000,000,000 personas},
  author={Ge, Tao and Chan, Xin and Wang, Xiaoyang and Yu, Dian and Mi, Haitao and Yu, Dong},
  journal={arXiv preprint arXiv:2406.20094},
  year={2024}
}

@article{li2015diversity,
  title={A diversity-promoting objective function for neural conversation models},
  author={Li, Jiwei and Galley, Michel and Brockett, Chris and Gao, Jianfeng and Dolan, Bill},
  journal={arXiv preprint arXiv:1510.03055},
  year={2015}
}

@inproceedings{li2024entropic,
  title={Entropic distribution matching for supervised fine-tuning of LLMs: Less overfitting and better diversity},
  author={Li, Ziniu and Chen, Congliang and Xu, Tian and Qin, Zeyu and Xiao, Jiancong and Sun, Ruoyu and Luo, Zhi-Quan},
  booktitle={NeurIPS 2024 Workshop on Fine-Tuning in Modern Machine Learning: Principles and Scalability},
  year={2024}
}

@inproceedings{li2025preserving,
  title={Preserving diversity in supervised fine-tuning of large language models},
  author={Li, Ziniu and Chen, Congliang and Xu, Tian and Qin, Zeyu and Xiao, Jiancong and Luo, Zhi-Quan and Sun, Ruoyu},
  booktitle={The Thirteenth International Conference on Learning Representations},
  year={2025}
}

@article{lanchantin2025diverse,
  title={Diverse Preference Optimization},
  author={Lanchantin, Jack and Chen, Angelica and Dhuliawala, Shehzaad and Yu, Ping and Weston, Jason and Sukhbaatar, Sainbayar and Kulikov, Ilia},
  journal={arXiv preprint arXiv:2501.18101},
  year={2025}
}

@article{chung2025modifying,
  title={Modifying Large Language Model Post-Training for Diverse Creative Writing},
  author={Chung, John Joon Young and Padmakumar, Vishakh and Roemmele, Melissa and Sun, Yuqian and Kreminski, Max},
  journal={arXiv preprint arXiv:2503.17126},
  year={2025}
}

@article{qin2025dive,
  title={DIVE: Diversified Iterative Self-Improvement},
  author={Qin, Yiwei and Liu, Yixiu and Liu, Pengfei},
  journal={arXiv preprint arXiv:2501.00747},
  year={2025}
}

@article{cideron2024diversity,
  title={Diversity-rewarded CFG distillation},
  author={Cideron, Geoffrey and Agostinelli, Andrea and Ferret, Johan and Girgin, Sertan and Elie, Romuald and Bachem, Olivier and Perrin, Sarah and Ram{\'e}, Alexandre},
  journal={arXiv preprint arXiv:2410.06084},
  year={2024}
}

@inproceedings{salkar2022self,
  title={Self-repetition in abstractive neural summarizers},
  author={Salkar, Nikita and Trikalinos, Thomas and Wallace, Byron C and Nenkova, Ani},
  booktitle={Proceedings of the conference. Association for Computational Linguistics. Meeting},
  volume={2022},
  pages={341},
  year={2022}
}

@article{shaib2024standardizing,
  title={Standardizing the measurement of text diversity: A tool and a comparative analysis of scores},
  author={Shaib, Chantal and Barrow, Joe and Sun, Jiuding and Siu, Alexa F and Wallace, Byron C and Nenkova, Ani},
  journal={arXiv preprint arXiv:2403.00553},
  year={2024}
}

@article{mccarthy2010mtld,
  title={MTLD, vocd-D, and HD-D: A validation study of sophisticated approaches to lexical diversity assessment},
  author={McCarthy, Philip M and Jarvis, Scott},
  journal={Behavior research methods},
  volume={42},
  number={2},
  pages={381--392},
  year={2010},
  publisher={Springer}
}

@article{covington2010cutting,
  title={Cutting the Gordian knot: The moving-average type--token ratio (MATTR)},
  author={Covington, Michael A and McFall, Joe D},
  journal={Journal of quantitative linguistics},
  volume={17},
  number={2},
  pages={94--100},
  year={2010},
  publisher={Taylor \& Francis}
}

@article{mass1972zusammenhang,
  title={{\"U}ber den zusammenhang zwischen wortschatzumfang und l{\"a}nge eines textes},
  author={Mass, Heinz-Dieter},
  journal={Zeitschrift f{\"u}r Literaturwissenschaft und Linguistik},
  volume={2},
  number={8},
  pages={73},
  year={1972},
  publisher={Vandenhoeck und Ruprecht}
}

@article{bellemare2024divergent,
  title={Divergent creativity in humans and large language models},
  author={Bellemare-Pepin, Antoine and Lespinasse, Fran{\c{c}}ois and Th{\"o}lke, Philipp and Harel, Yann and Mathewson, Kory and Olson, Jay A and Bengio, Yoshua and Jerbi, Karim},
  journal={arXiv preprint arXiv:2405.13012},
  year={2024}
}

@article{johnson2023divergent,
  title={Divergent semantic integration (DSI): Extracting creativity from narratives with distributional semantic modeling},
  author={Johnson, Dan R and Kaufman, James C and Baker, Brendan S and Patterson, John D and Barbot, Baptiste and Green, Adam E and van Hell, Janet and Kennedy, Evan and Sullivan, Grace F and Taylor, Christa L and others},
  journal={Behavior Research Methods},
  volume={55},
  number={7},
  pages={3726--3759},
  year={2023},
  publisher={Springer}
}

@article{deshpande2025penalty,
  title={A Penalty Goes a Long Way: Measuring Lexical Diversity in Synthetic Texts Under Prompt-Influenced Length Variations},
  author={Deshpande, Vijeta and Dasgupta, Ishita and Bhattacharya, Uttaran and Sarkhel, Somdeb and Mitra, Saayan and Rumshisky, Anna},
  journal={arXiv preprint arXiv:2507.15092},
  year={2025}
}

@article{gomez2023confederacy,
  title={A confederacy of models: A comprehensive evaluation of LLMs on creative writing},
  author={G{\'o}mez-Rodr{\'\i}guez, Carlos and Williams, Paul},
  journal={arXiv preprint arXiv:2310.08433},
  year={2023}
}

@article{chakrabarty2023creativity,
  title={Creativity support in the age of large language models: An empirical study involving emerging writers},
  author={Chakrabarty, Tuhin and Padmakumar, Vishakh and Brahman, Faeze and Muresan, Smaranda},
  journal={arXiv preprint arXiv:2309.12570},
  year={2023}
}

@article{chakrabarty2024can,
  title={Can AI writing be salvaged? Mitigating Idiosyncrasies and Improving Human-AI Alignment in the Writing Process through Edits},
  author={Chakrabarty, Tuhin and Laban, Philippe and Wu, Chien-Sheng},
  journal={arXiv preprint arXiv:2409.14509},
  year={2024}
}

@inproceedings{evans2016learning,
  title={Learning the preferences of ignorant, inconsistent agents},
  author={Evans, Owain and Stuhlm{\"u}ller, Andreas and Goodman, Noah},
  booktitle={Proceedings of the AAAI Conference on Artificial Intelligence},
  volume={30},
  number={1},
  year={2016}
}

@article{rafailov2023direct,
  title={Direct preference optimization: Your language model is secretly a reward model},
  author={Rafailov, Rafael and Sharma, Archit and Mitchell, Eric and Manning, Christopher D and Ermon, Stefano and Finn, Chelsea},
  journal={Advances in Neural Information Processing Systems},
  volume={36},
  pages={53728--53741},
  year={2023}
}

@article{hu2021lora,
  title={Lora: Low-rank adaptation of large language models. arXiv 2021},
  author={Hu, Edward J and Shen, Yelong and Wallis, Phillip and Allen-Zhu, Zeyuan and Li, Yuanzhi and Wang, Shean and Wang, Lu and Chen, Weizhu},
  journal={arXiv preprint arXiv:2106.09685},
  year={2021}
}

@article{dettmers2023qlora,
  title={Qlora: Efficient finetuning of quantized llms},
  author={Dettmers, Tim and Pagnoni, Artidoro and Holtzman, Ari and Zettlemoyer, Luke},
  journal={Advances in neural information processing systems},
  volume={36},
  pages={10088--10115},
  year={2023}
}

@article{tian2024toward,
  title={Toward self-improvement of llms via imagination, searching, and criticizing},
  author={Tian, Ye and Peng, Baolin and Song, Linfeng and Jin, Lifeng and Yu, Dian and Han, Lei and Mi, Haitao and Yu, Dong},
  journal={Advances in Neural Information Processing Systems},
  volume={37},
  pages={52723--52748},
  year={2024}
}

@inproceedings{lee2013pseudo,
  title={Pseudo-label: The simple and efficient semi-supervised learning method for deep neural networks},
  author={Lee, Dong-Hyun and others},
  booktitle={Workshop on challenges in representation learning, ICML},
  volume={3},
  number={2},
  pages={896},
  year={2013},
  organization={Atlanta}
}

@article{amini2025self,
  title={Self-training: A survey},
  author={Amini, Massih-Reza and Feofanov, Vasilii and Pauletto, Loic and Hadjadj, Lies and Devijver, Emilie and Maximov, Yury},
  journal={Neurocomputing},
  volume={616},
  pages={128904},
  year={2025},
  publisher={Elsevier}
}

@article{deshpande2023honey,
  title={Honey, I shrunk the language: Language model behavior at reduced scale},
  author={Deshpande, Vijeta and Pechi, Dan and Thatte, Shree and Lialin, Vladislav and Rumshisky, Anna},
  journal={arXiv preprint arXiv:2305.17266},
  year={2023}
}

@article{muckatira2024emergent,
  title={Emergent abilities in reduced-scale generative language models},
  author={Muckatira, Sherin and Deshpande, Vijeta and Lialin, Vladislav and Rumshisky, Anna},
  journal={arXiv preprint arXiv:2404.02204},
  year={2024}
}

@article{kaplan2020scaling,
  title={Scaling laws for neural language models},
  author={Kaplan, Jared and McCandlish, Sam and Henighan, Tom and Brown, Tom B and Chess, Benjamin and Child, Rewon and Gray, Scott and Radford, Alec and Wu, Jeffrey and Amodei, Dario},
  journal={arXiv preprint arXiv:2001.08361},
  year={2020}
}

@article{prabhakaran_thin_2014,
	title = {Thin slices of creativity: {Using} single-word utterances to assess creative cognition.},
	volume = {46},
	issn = {1554-3528(Electronic),1554-351X(Print)},
	doi = {10.3758/s13428-013-0401-7},
	number = {3},
	journal = {Behavior Research Methods},
	author = {Prabhakaran, Ranjani and Green, Adam E. and Gray, Jeremy R.},
	year = {2014},
	note = {Place: Germany
Publisher: Springer},
	pages = {641--659},
}

@article{guilford_structure_1956,
	title = {The structure of intellect.},
	volume = {53},
	issn = {1939-1455(Electronic),0033-2909(Print)},
	doi = {10.1037/h0040755},
	number = {4},
	journal = {Psychological Bulletin},
	author = {Guilford, J. P.},
	year = {1956},
	note = {Place: US
Publisher: American Psychological Association},
	pages = {267--293},
}

@article{olson_naming_2021,
	title = {Naming unrelated words predicts creativity.},
	volume = {118},
	copyright = {Copyright © 2021 the Author(s). Published by PNAS.},
	issn = {1091-6490 0027-8424},
	doi = {10.1073/pnas.2022340118},
	language = {eng},
	number = {25},
	journal = {Proceedings of the National Academy of Sciences of the United States of America},
	author = {Olson, Jay A. and Nahas, Johnny and Chmoulevitch, Denis and Cropper, Simon J. and Webb, Margaret E.},
	month = jun,
	year = {2021},
	pmid = {34140408},
	pmcid = {PMC8237676},
	note = {Place: United States},
}

@article{patterson2023multilingual,
  title={Multilingual semantic distance: Automatic verbal creativity assessment in many languages.},
  author={Patterson, John D and Merseal, Hannah M and Johnson, Dan R and Agnoli, Sergio and Baas, Matthijs and Baker, Brendan S and Barbot, Baptiste and Benedek, Mathias and Borhani, Khatereh and Chen, Qunlin and others},
  journal={Psychology of Aesthetics, Creativity, and the Arts},
  volume={17},
  number={4},
  pages={495},
  year={2023},
  publisher={Educational Publishing Foundation}
}

\appendix

\counterwithin{equation}{section}
\counterwithin{figure}{section}
\counterwithin{table}{section}

\section*{Appendix}

\section{Prompts}
This section provides the exact prompts used for data generation, model training, and model evaluation. 

\subsection{Data Generation Prompts}
\label{sec:app:prompts_data_gen}
The prompt used for generating the first response set from the model is as follows,
\begin{tcolorbox}[
    colback=gray!05,  
    colframe=black!50, 
    boxrule=0.5pt,     
    arc=2mm,           
    width=\linewidth,  
    leftrule=0.1mm,      
    rightrule=0.1mm,     
    bottomrule=0.1mm,    
    toprule=0.1mm,       
]

\texttt{\textbf{System Prompt:}}
\texttt{Task Description: For this task, you will write a very short story. You will be given 3 words, and write a story that includes all 3 words. Your story should be about 5 sentences long. Use your imagination and be creative when writing your story. But, also be sure your story makes sense.}

\texttt{\textbf{User Prompt:}}
\texttt{Write a short story that includes these three words: [THREE\_WORDS].}
\end{tcolorbox}

The prompt used for generating the second response set from the model is as follows,
\begin{tcolorbox}[
    colback=gray!05,  
    colframe=black!50, 
    boxrule=0.5pt,     
    arc=2mm,           
    width=\linewidth,  
    leftrule=0.1mm,      
    rightrule=0.1mm,     
    bottomrule=0.1mm,    
    toprule=0.1mm,       
]

\texttt{\textbf{System Prompt:}}
\texttt{Task Description: For this task, you will write a very short story. You will be given 3 words, and write a story that includes all 3 words. Your story should be about 5 sentences long. Use your imagination and be creative when writing your story. But, also be sure your story makes sense.}

\texttt{\textbf{User Prompt:}}
\texttt{Write a short story that includes these three words: [THREE\_WORDS].}

\texttt{\textbf{Assistant Prompt:}}
\texttt{[FIRST\_STORY]}

\texttt{\textbf{User Prompt:}}
\texttt{I do not like the previous story. Please rewrite the story in the most creative way. The new story:
- must be completely different from the previous story in: story plot and characters.
- must have a completely different start (do not use standard phrases like "Once upon", "As the", "In a", "In the" etc.).
- must be composed of exactly [FIRST\_STORY\_WORD\_COUNT] words.
Remember to use the three words: [THREE\_WORDS]}
\end{tcolorbox}

\subsection{Model Evaluation Prompts}
\label{sec:app:prompts_model_eval}

\paragraph{Divergent Association Task}
The prompt used for the Divergent Association Task (DAT) is as follows, 
\begin{tcolorbox}[
    colback=gray!05,  
    colframe=black!50, 
    boxrule=0.5pt,     
    arc=2mm,           
    width=\linewidth,  
    leftrule=0.1mm,      
    rightrule=0.1mm,     
    bottomrule=0.1mm,    
    toprule=0.1mm,       
]
\texttt{\textbf{System Prompt:}}
\texttt{Task description: Please generate 10 words that are as different from each other as possible, in all meanings and uses of the words. Rules: Only single words in English. Only nouns (e.g., things, objects, concepts). No proper nouns (e.g., no specific people or places). No specialized vocabulary (e.g., no technical terms). Think of the words on your own (e.g., do not just look at objects in your surroundings). Make a list of these 10 words, without any repetition. You must list each word with a number and a period. For example, "1. word-1, 2. word-2, etc." }

\texttt{\textbf{User Prompt:}}
\texttt{List 10 words that are as different from each other as possible: }
\end{tcolorbox}

\paragraph{Persona Generation Task (PGT)}
The prompt used for the Persona Generation Task (PGT) is as follows, 
\begin{tcolorbox}[
    colback=gray!05,  
    colframe=black!50, 
    boxrule=0.5pt,     
    arc=2mm,           
    width=\linewidth,  
    leftrule=0.1mm,      
    rightrule=0.1mm,     
    bottomrule=0.1mm,    
    toprule=0.1mm,       
]
\texttt{\textbf{System Prompt:}}
\texttt{Generate a random persona description with three characteristics.
Characteristics are:
- First Name
- The city of birth
- Current occupation 
Format the output strictly using JSON schema. Use ‘first\_name‘ for First Name, ‘city‘ for the city of birth, ‘occupation‘ for current occupation as corresponding JSON keys. The ordering of characteristics should be arbitrary in your answer.}
\end{tcolorbox}

\paragraph{Alternate Uses Task (AUT).}
The prompt used for the Alternate Uses Task (AUT) is as follows, 
\begin{tcolorbox}[
    colback=gray!05,  
    colframe=black!50, 
    boxrule=0.5pt,     
    arc=2mm,           
    width=\linewidth,  
    leftrule=0.1mm,      
    rightrule=0.1mm,     
    bottomrule=0.1mm,    
    toprule=0.1mm,       
]
\texttt{\textbf{System Prompt:}}
\texttt{Task Description: For this task, you'll be asked to come up with as many original and creative uses for objects as you can.
The goal is to come up with creative ideas, which are ideas that strike people as clever, unusual, interesting, uncommon, humorous, innovative, or different.
You must list each use with a number and a period. For example, "1. Use-1, 2. Use-2, 3. Use-3, etc.". You must provide exactly five (5) uses for each object.}

\texttt{\textbf{User Prompt:}}
\texttt{Object: [OBJECT], Uses: }
\end{tcolorbox}

The objects used for collecting the AUT responses are as follows, 
\begin{tcolorbox}[
    colback=gray!05,  
    colframe=black!50, 
    boxrule=0.5pt,     
    arc=2mm,           
    width=\linewidth,  
    leftrule=0.1mm,      
    rightrule=0.1mm,     
    bottomrule=0.1mm,    
    toprule=0.1mm,       
]
\texttt{"belt", "brick", "broom", "bucket", "candle", "clock", "comb", "knife", "lamp", "pencil", "pillow", "purse", "rope", "sock", "table"}
\end{tcolorbox}

\paragraph{Creative Writing Task (CWT).}

The three-word sets used in evaluating the model are as follows,
\begin{tcolorbox}[
    colback=gray!05,  
    colframe=black!50, 
    boxrule=0.5pt,     
    arc=2mm,           
    width=\linewidth,  
    leftrule=0.1mm,      
    rightrule=0.1mm,     
    bottomrule=0.1mm,    
    toprule=0.1mm,       
]
\texttt{
("stamp, letter, send"),
("petrol, diesel, pump"),
("statement, stealth, detect"),
("belief, faith, sing"),
("gloom, payment, exist"),
("organ, empire, comply"),
("year, week, embark"),
}
\end{tcolorbox}

\paragraph{Instruction for Human Evaluators}

The instruction provided to the human evaluators is as follows,
\begin{tcolorbox}[
    colback=gray!05,  
    colframe=black!50, 
    boxrule=0.5pt,     
    arc=2mm,           
    width=\linewidth,  
    leftrule=0.1mm,      
    rightrule=0.1mm,     
    bottomrule=0.1mm,    
    toprule=0.1mm,       
]
\texttt{
For each example, you will be given a [PROMPT] and two responses to that prompt: [RESPONSE-1] and [RESPONSE-2]. Your task is to compare the two responses based solely on text diversity—that is, the variety and novelty of vocabulary, sentence structure, and ideas. Do not consider other aspects such as correctness, relevance, or fluency.
Choose only one of the following options: A. [RESPONSE-1] is more diverse B. [RESPONSE-2] is more diverse C. Cannot decide
Respond with only the letter A, B, or C.
}
\end{tcolorbox}

\paragraph{Task Description for the LLM-as-a-Judge Evaluation}

The task description for the LLM-as-a-Judge evaluation is as follows,
\begin{tcolorbox}[
    colback=gray!05,  
    colframe=black!50, 
    boxrule=0.5pt,     
    arc=2mm,           
    width=\linewidth,  
    leftrule=0.1mm,      
    rightrule=0.1mm,     
    bottomrule=0.1mm,    
    toprule=0.1mm,       
]
\texttt{
You are an expert judge of text diversity.
For each example, you will be given a [PROMPT] and two responses to that prompt: [RESPONSE-1] and [RESPONSE-2]. Your task is to compare the two responses based solely on text diversity—that is, the variety and novelty of vocabulary, sentence structure, and ideas. Do not consider other aspects such as correctness, relevance, or fluency.
Choose only one of the following options: A. [RESPONSE-1] is more diverse B. [RESPONSE-2] is more diverse C. Cannot decide
Respond with only the letter A, B, or C.
}
\end{tcolorbox}

\section{Pilot Analysis for Sequential Prompting}
\label{sec:app:seq_prompting}

We conducted an exploratory analysis on $20,000$ short stories generated from Llama-3.1-8B and Olmo-2-7B models \cite{grattafiori2024llama, olmo20242}. The analysis was targeted at understanding the repeating patterns in the generated stories. With the help of the \textit{diversity} package in Python \cite{shaib2024standardizing}, we extract the top-5 repeating Part-Of-Speech (POS) bi-grams. We find that the most repeated bigram (\textit{IN DT}) occurs in over 15k stories (out of 20k) and $23\%$ of occurrences are present at the beginning of the generated story, refer to \cref{tab:app:repeating_patterns}. 

\begin{table*}[t]
\centering
\setlength{\tabcolsep}{4pt}
\small
\begin{tabular}{lccc}
\toprule
\textbf{POS Pattern} & \textbf{Example String} & 
\begin{tabular}[c]{@{}c@{}}\textbf{Present}\\ \textbf{(out of 20k)}\end{tabular} & 
\begin{tabular}[c]{@{}c@{}}\textbf{Present at}\\ \textbf{start (\%)}\end{tabular} \\
\midrule
IN DT  & \texttt{As a, In a, In the, At the, On the}        & $15{,}782$ & $23.30$ \\
DT JJ  & \texttt{a delicate, the rare, the main, the late}  & $11{,}418$ & $16.81$ \\
DT NN  & \texttt{an alley, a monarc, a spoon, a thicket}    & $18{,}472$ & $16.03$ \\
JJ NN  & \texttt{single silk, current king, ancient time}   & $9{,}335$  & $0.50$  \\
NN IN  & \texttt{hike in, group of, wave of, vendor to}     & $1{,}800$  & $0.45$  \\
\bottomrule
\end{tabular}

\vspace{2mm}

\caption{\textbf{Repeating bi-grams are more likely at the beginning.} We present the frequency of repeating POS bi-grams. \textit{IN DT} is the most frequent and commonly appears at the start of generated stories.}
\label{tab:app:repeating_patterns}
\end{table*}

Based on the findings, we conducted a sequential prompting experiment that elicits a more diverse response from the model by asking the model to avoid repeating phrases (refer to \cref{sec:app:prompts_data_gen} for exact prompts). We find that the diversity of the second response is, on average, higher than the first one.

\begin{table*}[t]
\centering
\small
\begin{tabular}{@{}lccc@{}}
\toprule
\textbf{Metric} & \textbf{First Story} & \textbf{Second Story} & \textbf{Increase in Diversity} \\
\midrule
TTR                 & $0.7112$  & $0.7469$  & $+0.0357$ \\
MAAS ($\downarrow$) & $0.1639$  & $0.1609$  & $+0.0031$ \\
HD-D                & $0.4143$  & $0.4202$  & $+0.0059$ \\
MTLD (MA-Bi)        & $13.9802$ & $14.3997$ & $+0.4195$ \\
MTLD (MA)           & $14.0778$ & $14.5063$ & $+0.4284$ \\
MTLD                & $14.2246$ & $14.6652$ & $+0.4406$ \\
MATTR               & $0.3810$  & $0.3867$  & $+0.0057$ \\
\bottomrule
\end{tabular}

\vspace{2mm}  

\caption{\textbf{Sequential prompting increases diversity.} We conducted a trial of sequential prompting on $20,000$ responses generated from Llama-8B and Olmo-7B models. The second story generated from the models has higher diversity. $\downarrow$: indicates that the lower values of MAAS index represent higher diversity.}
\label{tab:app:seq_prompt_increase_div}
\end{table*}

\section{Brief Description of ArmoRM score}
\label{sec:app:armo_rm}
ArmoRM uses a Llama-3-8B backbone frozen and attaches 19 “meter” heads that score crowd-rated qualities such as creativity, factuality, safety, and verbosity. A small prompt-conditioned gating network assigns non-negative weights to these meters and sums them, after explicitly removing any correlation with length to avoid verbosity bias. In practice the gate down-weights irrelevant meters: on creative-writing prompts $>70\%$ of the weight falls on creativity-related meters, $< 2\%$ on code-style. The model achieves $89\%$ pairwise accuracy on RewardBench, a benchmark whose preference labels are entirely human-annotated; this already reflects agreement with human judgements.

\section{Computational Savings Due to Lightweight Filtration}
\label{sec:app:computational_saving_d_ns_lite}

D-NS-Lite was explicitly designed with practical efficiency in mind. Computing entropy and ArmoRM scores for 400,000 generated sequences incurs two additional forward passes per example—one through the base language model and one through the reward model. With an average response length of 150 tokens, each pass processes approximately 60 million tokens. Assuming inference with a 7B-parameter model\footnote{We use Olmo-7B, LLaMA-8B, and ArmoRM, which has approximately 7.5B parameters.}, this results in an estimated $420 \times 10^{15}$ FLOPs per inference round based on the method outlined in \citet{kaplan2020scaling}, totaling around $840$ PFLOPs for both passes.
This computational cost—spent solely to measure diversity and quality—represents a significant overhead, roughly equivalent to or exceeding the cost of generating the data itself. As model sizes or dataset scales increase, the cost of metric computation will grow rapidly and can quickly become prohibitive. This motivates the need for low-cost alternatives to expensive metrics like entropy and reward model scores.
D-NS-Lite addresses this by replacing these metrics with lightweight lexical proxies such as TTR (Type-Token Ratio) and MAAS. These proxies entirely eliminate the need for the two additional inference rounds, resulting in substantial GPU computation savings. Although proxy metrics are computed on the CPU, their cost is negligible and can be handled efficiently even with modest hardware.
Notably, in sequential prompting setups (see \Cref{sec:sequential_prompting}), entropy computation for the first set of responses can be avoided, as these are already conditioned on the test-time prompt. However, if one wishes to use later-stage responses to construct the preference dataset, an additional forward pass would be required for each of those responses to match the intended test-time input distribution. By contrast, D-NS-Lite avoids this complication entirely, offering a significantly more efficient pipeline.

\section{Implementation of DivPO}
\label{sec:app:divpo_implementation}

\paragraph{Brief Description.} \citet{lanchantin2025diverse}, similar to our study, propose a data selection strategy for preference tuning to enhance response diversity in base models. Their approach involves two main steps for selecting a preference pair from $N$ responses to a fixed prompt. First, responses are ranked by a text quality score, with the top $\rho\%$ forming the pool of ``chosen'' candidates and the bottom $\rho\%$ forming the ``rejected'' pool. Second, each pool is sorted by a diversity metric. A single preference tuning pair is then formed by selecting the most diverse response from the ``chosen'' pool and the least diverse response from the ``rejected'' pool.

\paragraph{Hyperparameters.} The key hyperparameters in \citet{lanchantin2025diverse} are the text quality metric, the diversity metric, and the value of $\rho$. They experiment with rule-based rewards (e.g., JSON validity) and the ArmoRM reward model as quality metrics. In our study, we use ArmoRM exclusively. While they explore various $\rho$ values, we instead define the ``chosen'' pool as the top quartile (ArmoRM score $\geq 75^{\text{th}}$ percentile) and the ``rejected'' pool as the bottom quartile (ArmoRM score $\leq 25^{\text{th}}$ percentile). For diversity, they use log probabilities, whereas we use entropy.

\section{Hyperparameters for Preference Optimization}
\label{sec:app:hyperparameters}
We fine-tune the base model using the Direct Preference Optimization (DPO) objective \citep{rafailov2023direct}, with \(\beta = 0.1\) to control the divergence from the original policy. We use a peak learning rate of \(1 \times 10^{-5}\) with a cosine learning rate schedule, and a warm-up phase covering 10\% of the total training steps. All models are trained using LoRA adapters \citep{hu2021lora} with a rank \(r = 16\) and scaling factor \(\alpha = 16\), on a quantized 4-bit backbone model \citep{dettmers2023qlora}. We add the LoRA modules to \textit{query} and \textit{value} projection metrics of all transformer layers in the base model with a dropout of $5\%$.

\section{Reponse Length Distribution}
\label{sec:app:response_length}

We observe that the length distribution varies after fine-tuning the model. As presented in \cref{tab:app:post_training_response_length}, we observe that the average (and standard deviation) of response length reduces for DivPO and increases for our proposed methods (Diverse-NS and Diverse-NS-Lite). DivPO (inadvertently) teaches the model to generate shorter responses (refer to \cref{tab:post_filter_stats}). Despite maintaining comparable length for ``chosen'' and ``rejected'' samples in our methods (Diverse-NS and Diverse-NS-Lite), the model interestingly learns to generate longer responses. We suspect this shift is influenced by a skewed proportion of longer preference pairs, which may inadvertently bias the model toward generating longer responses.

\begin{table*}[ht]
\small
\centering
\begin{tabular}{lcccc}
\toprule
\textbf{Model} & \textbf{Base Model} & \textbf{DivPO \cite{lanchantin2025diverse}} & \textbf{Ours - D-NS-Lite} & \textbf{Ours - D-NS} \\
\midrule

Llama-8B    & $123.27 \pm \text{\scriptsize 18.14}$ & $111.24 \pm \text{\scriptsize 14.89}$ & $141.44 \pm \text{\scriptsize 37.26}$ & $139.47 \pm \text{\scriptsize 33.65}$ \\

Olmo-7B     & $73.63 \pm \text{\scriptsize 15.47}$  & $62.27 \pm \text{\scriptsize 12.88}$  & $81.37 \pm \text{\scriptsize 18.21}$  & $83.91 \pm \text{\scriptsize 17.93}$  \\
Olmo-13B    & $86.11 \pm \text{\scriptsize 13.96}$  & $72.20 \pm \text{\scriptsize 13.87}$  & $101.40 \pm \text{\scriptsize 17.64}$ & $100.60 \pm \text{\scriptsize 18.01}$ \\

\bottomrule
\end{tabular}

\vspace{2mm}  

\caption{\textbf{Change in the Response Length.} In this table, we present the average length of model-generated responses before and after the preference-tuning. The average values are calculated on 70 responses generated on the CWT evaluation prompts.}
\label{tab:app:post_training_response_length}
\end{table*}

\section{Results with Standard Deviation}
In this section, we report the results with the standard deviation values in \Cref{tab:app:all_combined_results}.

\begin{table*}[t]
\centering
\small
\begin{tabular}{llcccc}
\toprule

\textbf{Task} & \textbf{Metric} & \textbf{Base Model} & \textbf{DivPO} & \textbf{D-NS-Lite} & \textbf{D-NS} \\
\midrule

\multicolumn{6}{c}{\textbf{LLaMA-8B}} \\

\midrule
\midrule

DAT & DSI 
    & $0.7535 \pm \text{\scriptsize 0.07}$ 
    & $0.7545 \pm \text{\scriptsize 0.06}$ 
    & $0.7590 \pm \text{\scriptsize 0.07}$ 
    & $\mathbf{0.7640} \pm \text{\scriptsize 0.07}$ \\
DAT & Unique Words 
    & $0.4575$ & $0.4593$ & $0.4797$ & $\mathbf{0.4914}$ \\

PGT & Unique First Names 
    & $0.6500$ & $0.6100$ & $\mathbf{0.6900}$ & $\mathbf{0.6900}$ \\
PGT & Unique Cities 
    & $0.3300$ & $0.3100$ & $\mathbf{0.4700}$ & $0.4200$ \\
PGT & Unique Occupations 
    & $0.4100$ & $0.3900$ & $\mathbf{0.5100}$ & $\mathbf{0.4900}$ \\

AUT & DSI 
    & $0.8876 \pm \text{\scriptsize 0.02}$ 
    & $0.8837 \pm \text{\scriptsize 0.02}$ 
    & $0.8876 \pm \text{\scriptsize 0.02}$ 
    & $\mathbf{0.8878} \pm \text{\scriptsize 0.02}$ \\

CWT & DSI 
    & $0.8515 \pm \text{\scriptsize 0.01}$ 
    & $0.8521 \pm \text{\scriptsize 0.01}$ 
    & $0.8556 \pm \text{\scriptsize 0.01}$ 
    & $\mathbf{0.8581} \pm \text{\scriptsize 0.01}$ \\
CWT & ArmoRM Score 
    & $0.1451 \pm \text{\scriptsize 0.02}$ 
    & $\mathbf{0.1495} \pm \text{\scriptsize 0.01}$ 
    & $0.1369 \pm \text{\scriptsize 0.02}$ 
    & $0.1405 \pm \text{\scriptsize 0.02}$ \\
CWT & 4-gram div. POS
    & 0.4990
    & 0.4990
    & \textbf{0.5030}
    & 0.5000 \\
CWT & 4-gram div.
    & 2.8550
    & 2.9320
    & 2.9450
    & \textbf{2.9620} \\
CWT & Comp. Ratio.
    & 2.635
    & 2.546
    & 2.568
    & \textbf{2.530} \\

\midrule
\multicolumn{6}{c}{\textbf{OLMo-7B}} \\
\midrule
\midrule

DAT & DSI 
    & $0.7480 \pm \text{\scriptsize 0.09}$ 
    & $0.7509 \pm \text{\scriptsize 0.08}$ 
    & $\mathbf{0.7662} \pm \text{\scriptsize 0.08}$ 
    & $0.7639 \pm \text{\scriptsize 0.08}$ \\
DAT & Unique Words 
    & $0.6139$ & $0.6079$ & $\mathbf{0.6347}$ & $0.6327$ \\

PGT & Unique First Names 
    & $0.3300$ & $0.3300$ & $0.3300$ & $\mathbf{0.3400}$ \\
PGT & Unique Cities 
    & $\mathbf{0.3100}$ & $0.3000$ & $0.2700$ & $0.2700$ \\
PGT 
    & Unique Occupations & $0.5200$ & $0.5500$ & $\mathbf{0.6100}$ & $\mathbf{0.6100}$ \\

AUT & DSI 
    & $0.8836 \pm \text{\scriptsize 0.02}$ 
    & $0.8846 \pm \text{\scriptsize 0.02}$ 
    & $0.8852 \pm \text{\scriptsize 0.02}$ 
    & $\mathbf{0.8858} \pm \text{\scriptsize 0.02}$ \\

CWT & DSI 
    & $0.8499 \pm \text{\scriptsize 0.01}$ 
    & $0.8491 \pm \text{\scriptsize 0.01}$ 
    & $0.8548 \pm \text{\scriptsize 0.01}$ 
    & $\mathbf{0.8563} \pm \text{\scriptsize 0.01}$ \\
CWT & ArmoRM Score 
    & $0.1435 \pm \text{\scriptsize 0.02}$ 
    & $0.1441 \pm \text{\scriptsize 0.02}$ 
    & $0.1462 \pm \text{\scriptsize 0.01}$ 
    & $\mathbf{0.1464} \pm \text{\scriptsize 0.01}$ \\
CWT & 4-gram div. POS
    & 0.5720
    & \textbf{0.5770}
    & 0.5350
    & 0.5530 \\
CWT & 4-gram div.
    & 3.1270
    & 3.1690
    & \textbf{3.1750}
    & 3.1620 \\
CWT & Comp. Ratio.
    & 2.4460
    & 2.4160
    & \textbf{2.3850}
    & 2.3970 \\

\midrule
\multicolumn{6}{c}{\textbf{OLMo-13B}} \\
\midrule
\midrule

DAT & DSI 
    & $0.7233 \pm \text{\scriptsize 0.06}$ 
    & $0.7282 \pm \text{\scriptsize 0.07}$ 
    & $0.7320 \pm \text{\scriptsize 0.06}$ 
    & $\mathbf{0.7364} \pm \text{\scriptsize 0.06}$ \\
DAT & Unique Words 
    & $\mathbf{0.3421}$ & $0.3340$ & $0.3310$ & $0.3256$ \\
    
PGT & Unique First Names 
    & $0.4100$ & $0.4100$ & $0.4400$ & $\mathbf{0.4500}$ \\
PGT & Unique Cities 
    & $0.3500$ & $0.3500$ & $0.3700$ & $\mathbf{0.3900}$ \\
PGT & Unique Occupations 
    & $0.1900$ & $0.1900$ & $0.1900$ & $\mathbf{0.2000}$ \\

AUT & DSI 
    & $0.8943 \pm \text{\scriptsize 0.02}$ 
    & $0.8960 \pm \text{\scriptsize 0.02}$ 
    & $\mathbf{0.8974} \pm \text{\scriptsize 0.02}$ 
    & $0.8970 \pm \text{\scriptsize 0.02}$ \\

CWT & DSI 
    & $0.8557 \pm \text{\scriptsize 0.01}$ 
    & $0.8555 \pm \text{\scriptsize 0.01}$ 
    & $\mathbf{0.8616} \pm \text{\scriptsize 0.01}$ 
    & $0.8614 \pm \text{\scriptsize 0.01}$ \\
CWT & ArmoRM Score 
    & $0.1571 \pm \text{\scriptsize 0.01}$ 
    & $0.1589 \pm \text{\scriptsize 0.01}$ 
    & $0.1585 \pm \text{\scriptsize 0.01}$ 
    & $\mathbf{0.1590} \pm \text{\scriptsize 0.01}$ \\
CWT & 4-gram div. POS
    & 0.5210
    & \textbf{0.5229}
    & 0.5080
    & 0.4960 \\
CWT & 4-gram div.
    & 3.0820
    & 3.0770
    & 3.095
    & \textbf{3.1070} \\
CWT & Comp. Ratio.
    & 2.492
    & 2.512
    & 2.505
    & \textbf{2.480} \\

\bottomrule
\end{tabular}
\caption{\textbf{Diversity and Quality Evaluation.} We present the average ($\pm$ std. dev.) diversity (DSI or unique values) and quality (ArmoRM score) measurements for model responses collected on four creative generation tasks (Structured Gen.: DAT, PGT, Free-Form Gen.: AUT, CWT).}
\label{tab:app:all_combined_results}
\end{table*}

\section{$\Delta$DD-based Evaluation}
\label{sec:app:dd_eval}

Similar to the results presented \cref{fig:cwt:diversity_decile} for Olmo-7B, we present the results for Llama-8B and Olmo-13B in this section.

\begin{figure*}
    \centering
    \includegraphics[width=\linewidth]{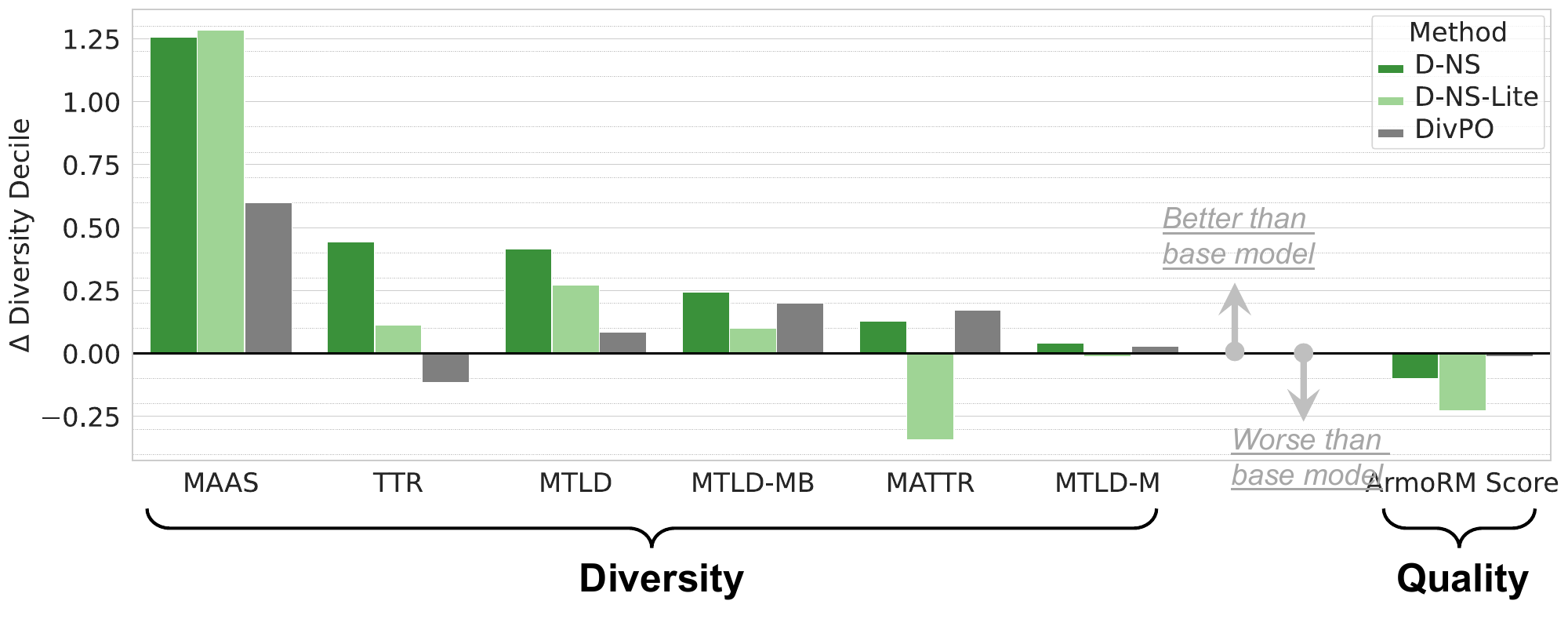}
    \includegraphics[width=\linewidth]{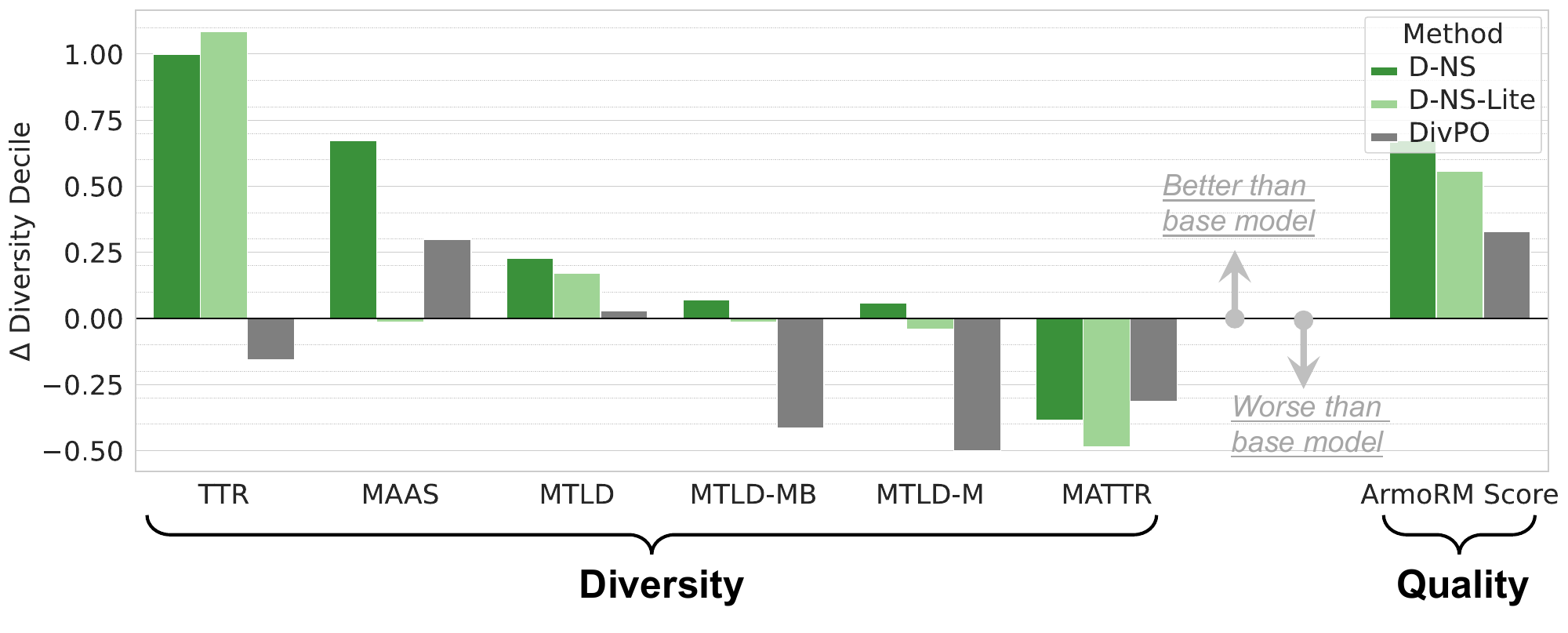}
    \caption{\textbf{Diversity and Quality Evaluation on CWT.}  
    This figure shows \(\Delta\)Diversity Decile (\(\Delta DD\)) values (y-axis) across various metrics (x-axis), computed from 70 CWT responses generated by the Llama-8B model (top-panel) and Olmo-13B (bottom panel). A value of zero represents base model performance; bars indicate improvements from preference-tuned models. 
    }
    \label{fig:app:cwt:diversity_decile}
\end{figure*}

\section{A Summary of Metrics}
We provide a concise summary of all metrics used in our evaluation setup in \Cref{tab:app:metrics}.

\begin{table*}[h]
\centering
\small
\begin{tabularx}{\textwidth}{@{}p{3cm} X X X@{}}
\toprule
\textbf{Metric} & \textbf{Definition} & \textbf{Trend Description (Trend)} & \textbf{Application} \\
\midrule
Entropy & Entropy of the token distribution in a response; measures unpredictability. & Higher values indicate greater lexical diversity (\(\uparrow\)). & Training-data filtering and diversity-bias analysis \\[3pt]
Type–Token Ratio (TTR) & Ratio of unique token types to total tokens. & Higher values indicate more lexical variety (\(\uparrow\)). & Lightweight filtering (D-NS-Lite), Calculation of Diversity Decile \\[3pt]
Moving-Average TTR (MATTR) & Moving-average of TTR over sliding windows; smooths variability. & Higher values indicate greater lexical diversity (\(\uparrow\)). & Correlation analysis, Calculation of Diversity Decile \\[3pt]
Measure of Textual Lexical Diversity (MTLD) & Average segment length until TTR falls below a threshold; longer segments imply more diversity. & Higher values indicate greater lexical diversity (\(\uparrow\)). & Correlation analysis, Calculation of Diversity Decile \\[3pt]
Moving-Average MTLD (MTLD-M) & Moving-average smoothing of MTLD to reduce variance. & Higher values indicate greater lexical diversity (\(\uparrow\)). & Correlation analysis, Calculation of Diversity Decile \\[3pt]
Bidirectional Moving-Average MTLD (MTLD-MB) & MTLD-M applied forward and backward for context-sensitive smoothing. & Higher values indicate greater lexical diversity (\(\uparrow\)). & Correlation analysis, Calculation of Diversity Decile \\[3pt]
MAAS & Proxy metric correlated with ArmoRM quality scores. & Higher values indicate stronger quality/diversity signal (\(\uparrow\)). & Lightweight filtering (D-NS-Lite), Calculation of Diversity Decile \\[3pt]
Hypergeometric Distribution Diversity (HD-D) & Probability-based measure of lexical diversity under a hypergeometric model. & Higher values indicate greater lexical diversity (\(\uparrow\)). & Correlation analysis \\[3pt]
ArmoRM score & Holistic quality score from a reward model. & Higher values indicate better fluency–diversity trade-off (\(\uparrow\)). & Quality evaluation (Creative Writing Task) and filtering, Calculation of Diversity Decile \\[3pt]
Divergent Semantic Integration (DSI) & Average semantic distance among items in a generated list. & Higher values indicate greater divergent thinking (\(\uparrow\)). & Diversity evaluation (Divergent Association Task, Creative Writing Task) \\[3pt]
Diversity Decile (DD) & Decile rank of a response’s diversity within its length group. & Higher decile indicates higher relative diversity after length normalization (\(\uparrow\)). & Length-normalized evaluation (Creative Writing Task) \\[3pt]
Change in Diversity Decile ($\Delta DD$) & Difference in DD before and after tuning; quantifies diversity gain. & Positive values indicate diversity gain; negative indicate loss (\(\uparrow\)/\(\downarrow\)). & Measuring tuning effect on diversity (Creative Writing Task) \\[3pt]
Semantic Distance (SD) & Average embedding-space distance between outputs; indicates semantic variety. & Higher values indicate greater semantic variety (\(\uparrow\)). & Diversity evaluation (Alternate Uses Task) \\
\bottomrule
\end{tabularx}
\caption{\textbf{Overview of diversity and quality metrics:} definitions, trend descriptions with arrows, and their applications including evaluation tasks.}
\label{tab:app:metrics}
\end{table*}

\section{LLM-as-a-Judge Evaluation on CWT}
\label{sec:app:llm_as_a_judge}


We extend our human evaluation setup to assess diversity gains using the LLM-as-a-Judge framework. Similar to the human study, we present a single prompt and two model responses to an LLM and ask it to select the more diverse output (see \Cref{sec:app:prompts_model_eval} for prompts). For each CWT prompt, we generate 10 responses per method, yielding 100 unique pairs for direct comparison. To ensure maximally distinguishable outputs, we select the 50 pairs with the lowest Jaccard similarity. Using 7 prompts, we construct 350 evaluation pairs per method comparison. Each comparison is evaluated by five proprietary LLMs: Claude-3.5-Haiku, Claude-Sonnet-4, GPT-4.1, GPT-4.1-mini, and GPT-4o-mini. We report the win percentages for D-NS and D-NS-Lite against the base model, DivPO, and each other in \Cref{tab:llm_as_a_judge_eval}.
The results align with human evaluation findings: D-NS outperforms both baselines in all but one case, and D-NS-Lite achieves similar improvements, also outperforming the baselines in all but one setting. The comparison between D-NS and D-NS-Lite is more competitive, with win rates hovering around $50\%$, slightly favoring D-NS.
\begin{table*}[t]
\centering
\begin{tabular}{@{}lccc@{}}
\toprule
\textbf{Judge} & \textbf{D-NS vs. Base} & \textbf{D-NS vs. DivPO} & \textbf{D-NS vs. D-NS-Lite} \\
\textbf{LLM} & D-NS Win $\%$ & D-NS Win $\%$ & D-NS Win $\%$ \\
\midrule
claude-3-5-haiku   & $54.86$ & $66.86$ & $47.43$ \\
claude-sonnet-4    & $52.86$ & $49.43$ & $54.57$ \\
gpt-4.1            & $74.86$ & $82.86$ & $54.86$ \\
gpt-4.1-mini       & $61.43$ & $78.86$ & $48.57$ \\
gpt-4o-mini        & $75.14$ & $82.86$ & $51.43$ \\
\midrule
\textbf{Judge} & \textbf{D-NS-Lite vs. Base} & \textbf{D-NS-Lite vs. DivPO} & \textbf{D-NS-Lite vs. D-NS} \\
\textbf{LLM} & D-NS-Lite Win $\%$ & D-NS-Lite Win $\%$ & D-NS-Lite Win $\%$ \\
\midrule
claude-3-5-haiku   & $58.57$ & $63.71$ & $52.00$ \\
claude-sonnet-4    & $48.86$ & $51.14$ & $45.14$ \\
gpt-4.1            & $77.43$ & $81.71$ & $45.14$ \\
gpt-4.1-mini       & $68.57$ & $78.00$ & $37.14$ \\
gpt-4o-mini        & $76.57$ & $81.43$ & $48.57$ \\
\bottomrule
\end{tabular}
\caption{\textbf{LLM-as-a-Judge Evaluation on CWT.} Pairwise LLM-as-a-Judge evaluation results comparing D-NS and D-NS-Lite against the base model, DivPO, and each other on the CWT task responses. Each cell presents the win percentage values. D-NS and D-NS-Lite achieve higher win rates against both baselines in all but two cases. The win rate values for D-NS versus D-NS-Lite comparison hover around $50\%$ while being slightly in favor of D-NS.}
\label{tab:llm_as_a_judge_eval}
\end{table*}

\section{Examples of Model Responses for CWT Prompts}
\label{sec:app:cwt_response_examples}
In this section, we provide a couple of examples of the model responses to CWT prompts, before and after the diversity tuning. Please refer to the \Cref{fig:app:cwt_response_examples}.

\begin{figure*}[t]
\centering
\begin{tcolorbox}[
    colback=gray!05,
    colframe=black!50,
    boxrule=0.5pt,
    arc=2mm,
    width=\textwidth, 
    leftrule=0.1mm,
    rightrule=0.1mm,
    bottomrule=0.1mm,
    toprule=0.1mm,
]
\texttt{\textbf{Prompt}: Write a creative story that includes the following words: belief, faith, sing.\\
\textbf{Base model response}: Once upon a time in a small town, belief and faith walked arm and arm. A man believed he could sing, despite having never uttered a note. One day, he decided to test the faith he had in his voice by entering a talent show. He sang with his heart, and to his amazement, the whole town believed his voice carried melody. His belief, supported by faith, made him that much more willing to sing, earning him first place and respect from all.\\
\textbf{Response after D-NS tuning}: In a quaint little village by a lush river, lived a community bound by belief and faith that a melody held magical powers. Every full moon, they gathered to sing under the shimmering silver light, believing their voices would protect them and bring prosperity to their lands. Faith, the backbone of their society, was unwavering despite the growing skepticism outside their village. That night, as they sang together, a gentle wave turned into a miraculous surge of bounty, renewing their faith and proving their belief in the power of song. The mesmerizing melody danced with the winds, a testament to their harmonious bond and shared beliefs.
}
\end{tcolorbox}

\begin{tcolorbox}[
    colback=gray!05,
    colframe=black!50,
    boxrule=0.5pt,
    arc=2mm,
    width=\textwidth, 
    leftrule=0.1mm,
    rightrule=0.1mm,
    bottomrule=0.1mm,
    toprule=0.1mm,
]
\texttt{\textbf{Prompt}: Write a creative story that includes the following words: year, week, embark.\\
\textbf{Response after DivPO tuning}: A new year brought new hope to the small town. Every resident embarked on a journey to make the year their best yet. One week in, Lucy decided to lose weight and improve her health. Her whole year proceeded as she planned, full of determination and hard work. By the end of the year, she had lost fifty pounds and was radiant with health.\\
\textbf{Response after D-NS tuning}: Every year, during the week of Christmas, a group of friends would embark on a magical adventure through the snowy town. Each year, their journey was marked by a whimsical event that seemed to defy reality, like talking animals or a melody that could heal the sick. This year, as winter approached, their anticipation was high. The week finally came, and, true to form, they discovered an ice-encased river that sang an ancient melody under the moonlight. Together, they ventured onto the ice, music echoing around them, and the frosty air filled their lungs, a reminder of magic's fleeting touch in our everyday year.
}
\end{tcolorbox}
\caption{\small \textbf{Model Responses Before and After Diversity Tuning.}}
\label{fig:app:cwt_response_examples}
\end{figure*}

\end{document}